# Using Model-Based Trees with Boosting to Fit Low-Order Functional ANOVA Models

Linwei Hu, Jie Chen, Arpita Mukherjee[1], and Vijayan N. Nair

Corporate Model Risk, Wells Fargo, USA

**Abstract**

Low-order functional ANOVA (fANOVA) models have been rediscovered in the machine learning (ML) community under the guise of inherently interpretable machine learning. Explainable Boosting Machines or EBM (Lou et al. 2013) and GAMI-Net (Yang et al. 2021) are two recently proposed ML algorithms for fitting functional main effects and second-order interactions. We propose a new algorithm, called GAMI-Tree, that is similar to EBM, but has a number of features that lead to better performance. It uses model-based trees as base learners and incorporates a new interaction filtering method that is better at capturing the underlying interactions. In addition, our iterative training method converges to a model with better predictive performance, and the embedded "purification" ensures that interactions are hierarchically orthogonal to main effects. The algorithm does not need extensive tuning, and our implementation is fast and efficient. We use simulated and real datasets to compare the performance and interpretability of GAMI-Tree with EBM and GAMI-Net.

## 1. Introduction

Interpretability of machine learning (ML) algorithms has been the subject of considerable discussion in recent years. Early approaches relied on post hoc techniques, including variable importance (Breiman 1996), partial dependence plots or PDPs (Friedman 2001), and H-statistics (Friedman and Popescu 2008). These are low-dimensional summaries of high-dimensional models with complex structure, and hence can be inadequate in capturing the full picture. A second approach for model interpretability is the use of surrogate models (or distillation techniques) that fit simpler models to extract information and explanations from the original complex models. Examples include: i) LIME (Ribeiro et al. 2016) based on linear models for local explanations; and ii) locally additive trees for global explanation (Hu et al. 2022).

A more recent direction is the use of ML algorithms to fit so-called inherently interpretable models that are extensions of the popular generalized additive models (GAMs) to incorporate common types of interactions. The rationale goes as follows. While there are applications (typically large-scale pattern recognition problems) where the use of very complex algorithms yields new results and insights, in many other areas, nonparametric models with lower-order interactions are sufficient in capturing the structure. This philosophy is a reversal of the trend towards fitting very complex ML models to squeeze out as much predictive performance as possible.

---

[1] Arpita Mukherjee contributed to this work when she was with Wells Fargo.

The additive index model (AIM), $g(x) = g_1(\boldsymbol{\beta}_1^T x) + g_2(\boldsymbol{\beta}_2^T x) + \ldots + g_K(\boldsymbol{\beta}_K^T x)$, is one way to generalize GAM to capture certain types of interactions. It was first proposed by Friedman and Stuetzle (1981) as an exploratory tool in the early days of nonparametric regression and was called projection pursuit. Vaughan et al. (2018) showed how one can use a restricted neural network to fit AIMs using gradient-based training. They referred to their method as explainable neural networks (xNNs).

Another class of models, based on low-order functional ANOVA (fANOVA), focuses on just the main effects (GAMs) and second-order interactions

$$g(x) = \sum_j g_j(x_j) + \sum_{j \neq k} g_{jk}(x_j, x_k). \qquad Eq~(1)$$

This class of fANOVA models are referred to as GA2M models in Lou et al. (2013). It should be noted that the concept of fANOVA has been around in the statistical literature for a long time, with some of the pioneering work done by Stone, Wahba, and her students (see Stone 1994; Gu 2013). The philosophy of approximating underlying models by low-order fANOVA structure of the form in Eq (1) is also well known. However, most of the available algorithms, based primarily on polynomial and smoothing splines, did not scale up to high-dimensions or large datasets. This is the gap that is being filled by recent papers that use ML architecture and their built-in fast algorithms to fit such models. Explainable boosting machine (EBM), in Lou et al. (2013), uses gradient boosting with piecewise constant trees to fit the GA2M models. GAMI-Net, developed by Yang et al. (2021), uses (restricted) neural network structures and the associated optimization techniques to fit the GA2M models.

EBM is a two-stage algorithm where the main effects and two-way interactions in Eq (1) are fitted in stages. Specifically: i) the main effect of each variable is modeled using small, piecewise-constant trees which split only on that single variable; and ii) the interaction effect of each pair is modeled using small trees (of depth 2) which split only on that same pair of variables. Within the main effect (or interaction) stage, the algorithm cycles through all variables (or pairs of variables) in a round-robin manner and iterates for several rounds. Since the total number of variable pairs can be large, an interaction filtering method, called FAST by the authors of EBM, is used to select the top interactions. Only those interactions are modeled in the second stage. In FAST, EBM fits a simple interaction model to the residuals (after removing the fitted main effects) for each pair of variables and ranks all pairs by the reduction in an appropriate metric for model error. The interaction model used in FAST is a simple approximation which divides the two-dimensional input space into four quadrants and fits a constant in each quadrant to estimate the functional interaction. Lou et al. (2013) justifies this approximation because fully building the interaction structure for each pair "is a very expensive operation".

GAMI-Net, by Yang et al. (2021), is also a multi-stage algorithm. It first uses GAM-Net (Agarwal et al. 2020), a specialized NN, to estimate the main effects. To impose sparsity, a pruning step is added at the end to remove variables/subnetworks with small contributions. Then the top interactions are then selected using the FAST algorithm from EBM (Lou et al., 2013) and are modeled using another specialized NN to capture interactions in the second stage. A pruning step is again added in the end to remove interactions with small contributions. Finally, all the important effects are collectively tuned in a final stage.

The goal of this paper is to propose a new model-based tree method, called GAMI-Tree, to fit the fANOVA model in Eq (1). Compared to EBM, our method has a few novelties, including a model-based tree base-learner, a more accurate interaction filtering method, a new iterative fitting method, and a purification step to enforce hierarchical orthogonality. The rest of the paper is organized as follows. Section 2 introduces the method, discusses the default hyper-parameter settings, model purification and model interpretation. The numerical implementation of the method is described in Section 3. The comparison with EBM is made in Section 4 to highlight the novelties of our algorithm. Section 5 describes simulation results comparing EBM, GAMI-Net and GAMI-Tree, followed by the real data analysis results in Section 6. Finally, we summarize our findings in Section 7.

## 2. GAMI-Tree

### 2.1 Methodology

We start by introducing the two specialized model-based trees (MBTs) that we use as base learners, followed by the boosting algorithm for fitting main-effect and interactions, a new interaction filtering algorithm, and finally the entire GAMI-Tree algorithm which iterates between the main-effect stage and the interaction-effect stage. M5 (Quinlan, 1992) is probably the earliest algorithm for fitting MBTs. See also LOTUS (Chan and Loh 2004) and MOB (Zeileis et al. 2008). EBM is also built on boosted trees, but it uses piecewise-constant trees as base learners. However, as Quinlan (1992) notes, MBTs are more parsimonious than piecewise-constant trees. In addition, the fitted response surface is less jumpy and provides a better approximation to the underlying model.

#### 2.1.1. Specialized Model-Based Trees (MBTs)

Two different and specialized MBTs are used in GAMI-Tree to capture main effects and interactions separately (see Figure 1).

a) The left panel is a tree for main effect, $T(x_j)$. It uses the same variable $x_j$ for both tree split and modeling inside the tree node. The model $f_v(x_j)$ inside node $v$ is chosen as a simple linear model, with L2 penalty to control overfitting. Repeated splits on the same variable will model inherent nonlinearities. Boosting allows it to further capture the complex nonlinear patterns.

b) The right panel shows the tree used to capture interaction effects, $T(x_j, x_k)$. The important $(x_j, x_k)$ pairs are identified in a separate step (discussed below). The algorithm first uses variable $x_k$ to split the tree, and then selects the other variable $x_j$ to model inside each node, allowing one to model this pairwise interactions. The model $f_v(x_j)$ inside node $v$ can be a simple linear model or a spline model after transforming $x_j$ with B-splines. The latter has more flexibility and can better capture complicated interaction patterns. But there is potential risk of overfitting. We apply L2 penalty to each model to control for overfitting.

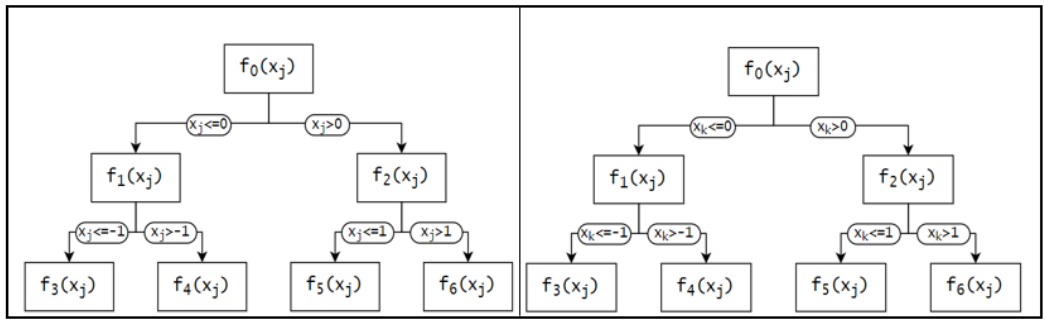

Figure 1. Main-effect tree $T(x_j)$ (left) and interaction-effect tree $T(x_j, x_k)$ (right)

## 2.1.2. Details of Boosting

We describe the boosting algorithm used in GAMI-Tree below to be self-contained. It follows the same framework that is used in xgboost, and the only difference is the base learner used. Let $x = (x_1, \ldots, x_j, \ldots, x_p)^T$ be the $p$-dimensional predictor vector, $g(x)$ be the model (to be fitted), and $y$ be the response. We consider both continuous and binary responses. For continuous response, we use the squared error loss function, $\ell(y, g(x)) = (y - g(x))^2$, and for binary response, we use log loss, $\ell(y, g(x)) = \log(1 + \exp(g(x))) - yg(x)$, where $g(x)$ is the log-odds. The goal is to minimize the mean loss $L = \frac{1}{n}\sum_{i=1}^{n} \ell(y_i, g(x_i))$ by boost it using model-based trees.

Specifically, let $M$ be the maximum number of boosting iterations and $g_0(x)$ be the initial value of $g(x)$ (eg., the overall mean or overall logit). For each boosting iteration $m, 1 \leq m \leq M$, we add a new MBT (multiplied by a learning rate $\lambda$) to the current model $g_{m-1}(x)$. We use the main-effect MBT $T_m(x_j)$ in the first stage and the interaction tree $T_m(x_j, x_k)$ in the second stage. For notational simplicity, we denote them below as $T_m^{(1)}(x)$ and $T_m^{(2)}(x)$, respectively. Following the approach in Xgboost, one can apply a second-order Taylor expansion to the loss function at each iteration to get:

$$\ell\left(y, g_{m-1} + T_m^{(s)}\right) \approx \ell(y, g_{m-1}) + \frac{\partial \ell(y, g_{m-1})}{\partial g_{m-1}} T_m^{(s)} + \frac{1}{2}\frac{\partial^2 \ell(y, g_{m-1})}{\partial g_{m-1}^2}\left[T_m^{(s)}\right]^2, \quad s = 1, 2.$$

Define for the $i$-th response, $G_{i,m-1} = \frac{\partial \ell(y_i, g_{m-1})}{\partial g_{m-1}}$ and $H_{i,m-1} = \frac{\partial^2 \ell(y_i, g_{m-1})}{\partial g_{m-1}^2}$. The total loss can then be approximated as

$$L \approx \frac{1}{n}\sum_{i=1}^{n} \ell(y_i, g_{m-1}(x_i)) + \frac{1}{n}\sum_{i=1}^{n} G_{i,m-1} T_m^{(s)}(x_i) + \frac{1}{2n}\sum_{i=1}^{n} H_{i,m-1}[T_m^{(s)}(x_i)]^2, \quad s = 1, 2.$$

Here $T_m^{(s)}(x_i)$ is the estimated prediction. Minimizing the approximate loss is equivalent to solving a weighted least square problem

$$\min_{T_m} \sum_{i=1}^{n} H_{i,m-1}\left(T_m^{(s)}(x_i) + \frac{G_{i,m-1}}{H_{i,m-1}}\right)^2.$$

Defining the "pseudo-response" $z_{i,m} = \frac{-G_{i,m-1}}{H_{i,m-1}}$ and weights $H_{i,m-1}$, we can find best model-based tree to minimize the sum of squared error[2]

$$SSE\left(T_m^{(s)}\right) = \sum_{i=1}^{n} H_{i,m-1}\left(z_{i,m} - T_m^{(s)}(x_i)\right)^2.$$

Table 1 and Table 2 summarize our algorithms for fitting the main effect stage and the interaction stage.

*Table 1. Algorithm 1 for Fitting Main Effects*

| **Algorithm 1: FitMain** |
|---|
| *Input: train, validation, initial $g_0(x)$* |
| for $m = 1$ to $M$ do: |

---

[2] When there is L2 penalty, GCV is used instead of SSE, which penalizes on effective degree of freedom.

| |
|---|
| calculate first and second order derivatives of the loss function: $G_{i,m-1}$ and $H_{i,m-1}$ |
| calculate pseudo-response $z_{i,m} = \frac{-G_{i,m-1}}{H_{i,m-1}}$ |
| for $j=1$ to $p$ do: |
|     fit a main effect tree $T_m^{(1)}(x_j)$ to $z_{i,m-1}$ using $x_j$, with weights $H_{i,m-1}$ |
|     calculate $SSE_j = \sum_{i=1}^{n} H_{i,m-1}\left(z_{i,m} - T_m^{(1)}(x_{ij})\right)^2$ |
| $j^* = \arg\min_j SSE_j$ |
| $g_m(x) = g_{m-1}(x) + \lambda T_m^{(1)}(x_{j^*})$ |
| calculate validation loss at $m$-th iteration $L_m = \frac{1}{n'}\sum_{i=1}^{n'} \ell(y_i, g(x_i))$, $n'$ is sample size |
| if $L_{m-d} < \min L_{m-d+1:m}$: # stop if no improvement in last $d$ iterations |
|     $g(x) = g_{m-d}(x)$ # the final $g(x)$ is rolled back to $d$ iteration earlier |
|     break |

*Table 2. Algorithm 2 for Fitting Interaction Effects*

| **Algorithm 2: FitInt** |
|---|
| *Input: train, validation, initial $g_0(x)$, top $q$ interaction pairs $Q$* |
| for $m = 1$ to $M$ do: |
|     calculate first and second order derivatives of the loss function: $G_{i,m-1}$ and $H_{i,m-1}$ |
|     calculate pseudo-response $z_{i,m} = \frac{-G_{i,m-1}}{H_{i,m-1}}$ |
|     for $(x_j, x_k)$ in $Q$ do: |
|         fit an interaction effect tree $T_m^{(2)}(x_j, x_k)$ to $z_{i,m}$, with weights $H_i$ |
|         calculate $SSE_{jk} = \sum_{i=1}^{n} H_{i,m-1}\left(z_{i,m} - T_m^{(2)}(x_{ij}, x_{ik})\right)^2$ |
|     $j^*, k^* = \arg\min_{j,k} SSE_{jk}$ |
|     $g_m(x) = g_m(x) + \lambda T_m^{(2)}(x_{j^*}, x_{k^*})$ |
|     calculate validation loss at $m$-th iteration $L_m = \frac{1}{n'}\sum_{i=1}^{n'} \ell(y_i, g(x_i))$, $n'$ is sample size |
|     if $L_{m-d} < \min L_{m-d+1:m}$: # stop if no improvement in last $d$ iterations |
|         $g(x) = g_{m-d}(x)$ # the final $g(x)$ is rolled back to $d$ iteration earlier |
|         break |

### 2.1.3. New Interaction Detection Algorithm

To select the top $q$ interaction pairs, we propose a new interaction filtering method. We first calculate the pseudo-residuals after fitting the main effects. Then for each pair of variables $(x_j, x_k)$, we fit two interaction trees to the pseudo-residuals, $T^{(2)}(x_j, x_k)$ and $T^{(2)}(x_k, x_j)$, by alternating their roles as modeling variable or splitting variable. By default, we use trees with maximum depth 2 and use linear B-splines with 5 knots (including two boundary knots) to transform the modeling variable. We use the smaller SSE of the two fitted interaction trees to measure the explain-ability power of the pair. Then we rank all pairs to select the top $q$ pairs and output $2q$ (modeling variable, splitting variable) combinations for the selected top $q$ pairs. See Table 3 for the detailed algorithm.

*Table 3. Algorithm 3 for Interaction Filtering*

| **Algorithm 3: FilterInt** |
|---|
| *Input: train, $g(x)$* |
| calculate first and second order derivatives of the loss function: $G_i$ and $H_i$ |
| calculate pseudo-response $z_i = \frac{-G_i}{H_i}$ |
| for $j = 1$ to $p - 1$ do: |
|    for $k = j + 1$ to $p$ do: |
|       fit an interaction effect tree $T^{(2)}(x_j, x_k)$ to $z_i$, with weights $H_i$ |
|       calculate $SSE_{jk} = \sum_{i=1}^n H_i \left(z_i - T^{(2)}(x_{ij}, x_{ik})\right)^2$ |
|       fit an interaction effect tree $T^{(2)}(x_k, x_j)$ to $z_i$, with weights $H_i$ |
|       calculate $SSE_{kj} = \sum_{i=1}^n H_i \left(z_i - T^{(2)}(x_{ik}, x_{ij})\right)^2$ |
|    rank all pairs by $\min(SSE_{jk}, SSE_{kj})$ in ascending order # choose the smaller SSE |
|    select top $q$ pairs $\{(j_1, k_1), \dots, (j_q, k_q)\}$ |
|    output $Q = \{(j_1, k_1), \dots, (j_q, k_q)\} \cup \{(k_1, j_1), \dots, (k_q, j_q)\}$ |

### 2.1.4. Description of Full GAMI-Tree Algorithm

At this point, we have all the building blocks to describe the full algorithm (see Table 4). We first initialize $g(x)$ to be the overall mean for continuous response or logit of overall probability for binary response. Then we repeatedly call the FitMain, FilterInt and FitInt functions (in that order) until the early stop iterations for the main effect stage ($M_{main\_stop}$) and interaction stage ($M_{int\_stop}$) are both 0, or the maximum number of iteration rounds ($R$) has been reached. Note the FitMain and FitInt functions use the output model from the previous stage as initial value $g_0(x)$, to continue adding main-effects/interactions to it. Therefore, it is another layer of boosting. The benefit of this additional layer of boosting will be illustrated in Section 5.

*Table 4. GAMI-Tree Algorithm*

| **Algorithm 4: GAMI-Tree** |
|---|
| *Input: train, validation* |
| Initialize $g(x) = g_0(x)$ to be overall mean (continuous) or overall logit (binary) |
| for $r = 1$ to $R$ do: |
|    FitMain(train, validation, $g(x)$), record early stopping rounds $M_{main\_stop}$ |
|    $Q$=FilterInt(train, $g(x)$) |
|    FitInt(train, validation, $g(x)$, $Q$), record early stopping rounds $M_{int\_stop}$ |
|    if $M_{main\_stop} == 0$ and $M_{int\_stop} == 0$: |
|       break # stop if both main and interaction stages stop with 0 iterations |

## 2.2 Hyper-parameter Tuning

We describe the key parameters and good default values. It turns out that GAMI-Tree requires relatively little tuning as there are good default values. We found that these default settings produce good models without extensive tuning.

- $M$: number of maximum boosting iterations in FitMain and FitInt function. As with xgboost, user only needs to set a large value and early stopping is used internally to find the best number of iterations. We set default value as 1000.
- max_depth: maximum depth in main-effect tree and interaction-tree. Shallower trees are preferred in boosting framework, as it has less overfitting issue compared to deeper trees. However, this will lead to more boosting iterations. Based on our experience in extensive simulations, we propose a default value of 2 for continuous response and 1 for binary response (as the pseudo-response in binary case is noisier).
- Learning rate $\lambda$: small learning rate is preferred in boosting framework, but it requires more boosting iterations to converge. We set a default value of 0.2, and users can reduce it to 0.1 or smaller if the dataset is small and noisy.
- nknots: number of linear B-spline transformation knots for the modeling variable when fitting *interaction-tree*. This gives more flexibility to the tree to capture complicated interaction patterns, but the number of knots needs to be small, so it does not overfit. We use a default value of 5, and pick 5 quantile knots (0, 25, 50, 75, and 100 percentiles). A special case of 2 knots means only the two boundary knots are included, hence it is equivalent to being linear.
- $R$: number of rounds of main effect stage and interaction stages. The benefit of adding a second round will be shown in Section 5. Usually the number of early stopping iterations ($M_{main\_stop}$ or $M_{int\_stop}$) in later round is small, so additional rounds does not add too much computational burden. We set a default value of $R = 5$ and use early stopping.
- npairs $q$: number of top interaction pairs to select in FilterInt in each round. We set the default value for $q$ as 10. Smaller values can be used without worrying about missing interactions since missed interactions in one round can be picked up in the next round.
- alpha: This is the L2 regularization parameter when fitting linear/spline regression models in each tree node. We use a default grid of penalty parameters from $\exp(-8)$ to $\exp(0)$. The best one is selected by GCV criterion. However, we found that sometimes the chosen penalty is not strong enough, so the algorithm includes a direct way to control overfitting using max_coef below.
- max_coef: maximum allowed coefficient value when fitting ridge regression model in each tree node. This will drop small L2 penalty parameters which produce normalized coefficient larger than max_coef, and choose only the best penalty from the remaining ones. Here normalized coefficient is defined by coefficient value times its standard deviation. We set default as 1, but user can also set it smaller to add more regularization.

## 2.3 Purifying Main Effects and Interactions

Even though the main effects and interactions are estimated separately, they may not satisfy the hierarchical orthogonality constraints $\int g_{jk}(x_j, x_k) g_\ell(x_\ell) w_{jk}(x_j, x_k) dx_j dx_k = 0$, $\ell = j, k$ where $w_{jk}(x_j, x_k)$ is the marginal density function of $x_j$ and $x_k$ (see, for example, Hooker 2007). However, this orthogonality is important for interpretation purposes. Therefore, we go through a "purification" process to enforce this condition.

Specifically, for a given $x_j$, we sum all main-effect trees $T_m^{(1)}(x_j)$ to form the initial estimate of $g_j(x_j)$. Similarly, for a given pair $(x_j, x_k)$, we sum all interaction effect trees $T_m^{(2)}(x_j, x_k)$ and $T_m^{(2)}(x_k, x_j)$ to form initial estimate of $g_{jk}(x_j, x_k)$. We obtain the pseudo-predictions on the training data as $\hat{y}_i^{jk} = g_{jk}(x_{ij}, x_{ik}), i = 1, \ldots, n$. Then we fit an additive model $h_j(x_j) + h_k(x_k)$ to $\hat{y}_i^{jk}$'s using B-spline transformed $x_j$ and $x_k$ variables (other ways of fitting additive models can also be used). Then, orthogonalized interactions are $\tilde{g}_{jk}(x_j, x_k) = g_{jk}(x_j, x_k) - h_j(x_j) - h_k(x_k)$. The subtracted main effects are added to $g_j(x_j)$ and $g_k(x_k)$: $\tilde{g}_j(x_j) = g_j(x_j) + h_j(x_j)$ and $\tilde{g}_k(x_k) = g_k(x_k) + h_k(x_k)$. With this approach, the final orthogonalized interaction and main effects are hierarchically orthogonal. A proof is provided in Appendix A, and we have also empirically verified this in all our studies. Other approaches of purification can also be used. For example, Lengerich (2020) proposed a method which iteratively removes the means projected onto a lower dimension covariate space, until all the lower dimensional means are 0. This is similar to what we are doing, except we estimate all the lower dimension means simultaneously instead of iteratively.

## 2.4 Measuring the Importance of Main Effects and Interactions

We can now measure the importance of each main effect and interaction through $M_j = std(\tilde{g}_j(x_j))$ and $I_{jk} = std(\tilde{g}_{jk}(x_j, x_k))$ respectively. Note these importance scores are different from the permutation-based importance, as the latter combine main effects and interactions into an overall importance score. By estimating them separately, we can identify the contributions of each, rank them by the order of importance, and visualize the effects by plotting the function on a grid. Without orthogonalization, part of the main effects $g_j(x_j)$ or $g_k(x_k)$ might be absorbed into $g_{jk}(x_j, x_k)$, creating false interaction or distorting the size of interaction effects.

GAMI-Net implements orthogonality by clarity regularization. Theoretically, this can be set as high as possible to enforce orthogonality, but a large value could impact model performance. So, in our comparisons below, we chose a relatively small value (0.1), but this means main-effects and interactions may not be exactly orthogonal, creating some difference between GAMI-Net and GAMI-Tree.

## 3. Efficient Numerical Implementation

Constructing model-based tree is known to be computationally expensive, because many linear models need to be fitted and evaluated in order to determine the best tree split. What is worse, GAMI-Tree requires fitting hundreds or even thousands of model-based trees in the boosting process. To address the computation obstacle, we have made an efficient implementation which reduces the computation by reusing intermediate results and utilizes high performance computational tools like multi-processing and Cython to speed it up.

First, to fit each model-based tree (either the main-effect tree or interaction-effect tree), we use the efficient algorithm proposed in Hu et al. (2020). Briefly, we bin the splitting variable and calculate the gram matrices, $\boldsymbol{X}^T\boldsymbol{X}, \boldsymbol{X}^T\boldsymbol{z}$, for each bin as intermediate results. Then in each tree node, we only need to find all the bins which fall into that node and sum over the corresponding binned gram matrices to obtain the gram matrix, instead of computing it from scratch. This reduces the computation cost tremendously when sample size $n$ is large since the most computation cost is in calculating the gram matrices in our applications ($n \gg p$). Moreover, only the pseudo-response $\boldsymbol{z}$ changes while the predictors stay fixed from iteration to iteration, so we can reuse the gram matrices for $X^T X$ and only updating the gram matrices for $X^T z$. This is fast because $z$ is one-dimensional.

In addition, we use high performance computational tools to speed it up. The gram calculation, loss evaluation function, prediction function and solver for the ridge regression are all written in Numba or Cython, which is compiled into C code and has the speed of C. These functions are further parallelized by joblib and openmp. So, the final algorithm we have is highly optimized and parallelized.

Table 5 shows the timing for fitting a GAMI-Tree model to a simulated binary response data with $n =$100K/1M/10M observations and $p = 50$ variables. The data is divided into 70% training and 30% validation, and a GAMI-Tree model with a particular hyper-parameter configuration (max_depth=2, ntrees=100, npairs=10, nknots=6, nrounds=1) is fitted to obtain the timing. Since the timing of GAMI-Tree model varies depending on how many rounds and number of trees are fitted, it is useful to show the time for each tree iteration. Table 5 shows the average time per tree in main-effect stage and interaction stage, time for interaction filtering and total fitting and prediction time. For small data with 100K observations, it is very fast, takes less than 0.1 seconds to fit one tree. For medium data with 1M rows, it takes 0.1-0.2 seconds to fit one tree. For large data with 10M rows, it takes less than 0.7 seconds to fit one tree for nthreads=20 and less than 1.2 seconds for nthreads=10. Regarding interaction filtering, it takes only 2 second to filter all 2500 pairs of variables for the 100K data, 6-9 seconds for the 1M data and 52-75 seconds for the entire 10M data. A lot of times, we find using just a 1M subsample to filter interactions is sufficient (since the interaction model is only a two-variable model), but even with the entire 10M data, the filter speed is still acceptable. In terms of total fitting time, for the largest 10M data, a typical GAMI-Tree with a few hundred trees for both main-effect stage and interaction stage can be done around 10 minutes. The prediction speed is even faster, taking less than 10 seconds for the 10M data.

*Table 5. Computational Times for GAMI-Tree*

| n | p | nthreads | main-stage (s/tree) | int-filter (s) | int-stage (s/tree) | total fit (s) | predict (s) |
|---|---|---|---|---|---|---|---|
| 100K | 50 | 10 | 0.08 | 2.2 | 0.06 | 18 | 0.15 |
| 100K | 50 | 20 | 0.08 | 2.0 | 0.06 | 18 | 0.22 |
| 1M | 50 | 10 | 0.18 | 9.0 | 0.12 | 44 | 1 |
| 1M | 50 | 20 | 0.15 | 6.0 | 0.1 | 36 | 1.1 |
| 10M | 50 | 10 | 1.20 | 75 | 0.82 | 312 | 9.5 |
| 10M | 50 | 20 | 0.70 | 52 | 0.53 | 224 | 6.5 |

## 4. Comparison with EBM

Both GAMI-Tree and EBM are tree-based algorithms, and they share several similarities: estimating main effects and interactions in separate stage, interaction filtering, and model-fitting in an additive way using simple base learners. However, there are some key differences as described below:

1. We use MBTs as base learners in fitting main effects and second order interactions. As noted before, MBTs are more flexible and require fewer splits and fewer number of trees to capture a complex function. In general, they lead to less overfitting and hence they have better generalization performance. We observed this in another study (Aramideh et al. 2022)
2. We propose a new interaction filtering method using MBTs. Even though the simple 4-quadrant model used in FAST works well in general, model-based tree can capture interaction pattern better and rank the interaction effects more accurately in some cases (see results in Section 5).
3. We use an iterative fitting method to fit the main effects and interactions, instead of the two-stage fitting method used in EBM. This has two advantages listed below, and they lead to performance improvement if we iterate.
    - When main effects and interaction terms are not orthogonal, fitting main effects and interaction terms cannot be done in the naïve two-stage way. As an analogy, if we think of the main effects and interaction terms as two correlated predictors $x_1, x_2$ (but not perfectly collinear), we cannot just fit $x_1$ first and then fit $x_2$ using the residuals; instead, we need to iteratively fit one predictor at a time until convergence (or fit the two simultaneously). Otherwise, we will find bias and worse model fit. We will demonstrate this in Section 5.
    - Some weaker interaction terms may be missed in the initial round of filtering. By iterating, we can capture the missed ones in the subsequent iterations. Therefore, it is better at capturing all true interactions. We will also demonstrate this in Section 5.
4. Instead of the round-robin training method used in EBM which cycles through all variables (or interaction pairs) at each round, we only choose the most important variable (or interaction pair) to model at each iteration. Thus, it overfits less on the non-significant variables (interaction pairs).
5. Finally, we have implemented a purification step, so the main effects and interactions are hierarchically orthogonal based on the functional-ANOVA concept in Hooker (2007). This allows us to measure the importance of main-effects and interactions more accurately.

## 5. Simulation Results

We used several stimulation cases to compare xgboost, GAMI-Tree, GAMI-Net, and EBM in terms of model performance and model interpretation. In addition, we also consider GAMI-Tree-1 which is just the first round of the fitted GAMI-Tree. We will show the benefit of iterating between main-effect stage and interaction stage by comparing GAMI-Tree-1 with GAMI-Tree.

### 5.1 Simulation Set Up

We considered the following four models:

1. $g(x) = \sum_{j=1}^{5} x_j + \sum_{j=6}^{8} 0.5 x_j^2 + \sum_{j=9}^{10} x_j (x_j > 0) + \sum_{j=1}^{10} \sum_{k=j+1}^{10} 0.2 x_j x_k$
2. $g(x) = \sum_{j=1}^{5} x_j + \sum_{j=6}^{8} 0.5 x_j^2 + \sum_{j=9}^{10} x_j (x_j > 0) + 0.25 x_1 x_2 + 0.25 x_1 x_3^2 + 0.25 x_4^2 x_5^2 + \exp(x_4 x_6 / 3) + x_5 x_6 (x_5 > 0)(x_6 > 0) + clip(x_7 + x_8, -1, 0) + clip(x_7 x_9, -1, 1) + (x_8 > 0)(x_9 > 0)$.

   Here $clip(x, a, b)$ is the cap and floor function. It caps the value of $x$ at $b$ and floors it at $a$.

3. $g(x) = \sum_{j=1}^{5} x_j + \sum_{j=6}^{8} 0.5 x_j^2 + \sum_{j=9}^{10} x_j (x_j > 0) + 0.25 x_1^2 x_2^2 + 2(x_3 - 0.5)_+ (x_4 - 0.5)_+ + 0.5 \sin(\pi x_5) \sin(\pi x_6) + 0.5 \sin(\pi (x_7 + x_8))$
4. $g(x) = \sum_{j=1}^{5} x_j + \sum_{j=6}^{8} 0.5 x_j^2 + \sum_{j=9}^{10} x_j (x_j > 0) + x_1 x_2 + x_1 x_3 + x_2 x_3 + 0.5 x_1 x_2 x_3 + x_4 x_5 + x_4 x_6 + x_5 x_6 + 0.5 (x_4 > 0) x_5 x_6$

Model 1 contains a total of 45 interactions, and we wanted to see if all of them can be captured. For model 2, we considered eight different forms of interactions. Model 3 includes the oscillating sine functions, which would be hard to capture by the 4-quandrat approximation used in FAST. Model 4 contains two 3-way interactions; we included then here to assess the performance of the GA2M models. In practice, they will capture only the projection of 3-order interactions into one and two-dimensions.

For each model form, we simulated 20 predictors $x_1 \sim x_{20}$ from multivariate Gaussian distribution with mean 0, variance 1 and equal correlation $\rho$. Only the first 10 variables, $x_1 \sim x_{10}$, are used in the model, and the rest are not part of the model although they will be relevant when $\rho > 0$ (*redundant* variables). We then simulated 10 additional variables $x_{21} \sim x_{30}$ which were independent of the first 20 variables (*irrelevant* variables). They were also simulated from multivariate Gaussian distribution with mean 0, variance 1 and equal correlation $\rho$. So there were 30 predictors in total. To avoid potential outliers in $x$ to be too influential, we truncated all predictors to be within the interval $[-2.5, 2.5]$.

The response was simulated as $y = g(x) + \epsilon$, $\epsilon \sim N(0, 0.5^2)$ for continuous case and as $Bernoulli(p(x))$ for binary case, where $p(x) = \frac{e^{\beta_0 + g(x)}}{1 + e^{\beta_0 + g(x)}}$ and the intercept $\beta_0$ was carefully chosen to have balanced classes. We considered two correlation levels $\rho = 0, 0.5$. For each model form and correlation level, we simulated datasets with two different sample sizes, 50K and 500K. Each dataset was divided into train, validation, and testing sets, with 50%, 25% and 25% sample sizes, respectively. We used training set and validation set to train and tune four models (xgboost, EBM, GAMI-Tree, and GAMI-Net) and evaluated the predictive performance on the test set. Below are the tuning settings.

- For EBM, we tune max_bins, max_interaction_bins and learning rate and fix the number of interaction pairs to be 45 for Model 1 and 10 for the other models. We use random search with a total number of 12 trials.
- For GAMI-Tree, we use default settings mentioned in Section 2.2. The only exception is we set npairs as 45 (instead of default 10) for Model 1.
- For GAMI-Net, we use a subnet architecture of 5 layers, each with 40 neurons. Number of epochs is set as 200, learning rate is set as 0.0001, batch size is set as 1000, number of interactions is 45 for Model 1 and 10 for the other models, and clarity penalty is set as 0.1. These settings were decided on after discussions with the authors of the GAMI-Net paper.
- For xgboost, we tune maximum depth and learning rate using grid search, and we use early stopping for the number of boosting rounds.

## 5.2 Simulation Results
### 5.2.1 Continuous Case

The training and testing mse for all models are listed in Table 6. From these results, we can conclude that:

- GAMI-Tree outperforms xgboost for all cases except for Model 4, $\rho = 0$. This is not surprising because Model 4 has 3-way interactions which are not captured entirely by GA2M models. However, when correlation increases, the 3-way interaction can be better approximated by lower order effects (in the extreme case when the correlation is 1, it becomes a main effect), and GAMI-Tree outperforms.
- GAMI-Tree and GAMI-Tree-1 are similar for uncorrelated case, but GAMI-Tree significantly outperforms GAMI-Tree-1 for correlated case except for Model 3, and they both outperform EBM in all cases. This shows for correlated case, the iterative training used in GAMI-Tree helps in model performance.
- GAMI-Tree has similar performance as GAMI-NET in most cases, except for Model 1, 50K, rho=0.5, Model 2, rho=0.5 and Model 3. For the first case, GAMI-NET has 10% smaller MSE. This is likely due to neural networks being better at capturing such linear interaction effects and are smoother. As sample size increases to 500K, this advantage becomes marginal. For Model 2, rho=0.5 and Model 3, GAMI-Tree is better. As we will see later, this is because the FAST interaction filtering method (used in both EBM and GAMI-Net) misses some true interactions terms.
- GAMI-Net has smaller train/test mse gap than all other models. This is known in the literature as NNs are smooth models and overfit less. Among the others, GAMI-Tree overfits less than EBM and xgboost.

*Table 6. Train and test mse for continuous simulation cases*

|  | N | rho | xgboost | | GAMI-Net | | EBM | | GAMI-Tree | | GAMI-Tree-1 | |
|---|---|---|---|---|---|---|---|---|---|---|---|---|
|  |  |  | train | test | train | test | train | test | train | test | train | test |
| Model 1 | 50K | 0 | 0.170 | 0.480 | 0.244 | 0.279 | 0.187 | 0.318 | 0.234 | 0.288 | 0.236 | 0.287 |
| Model 1 | 50K | 0.5 | 0.346 | 0.629 | 0.244 | 0.287 | 0.604 | 0.900 | 0.243 | 0.317 | 0.502 | 0.609 |
| Model 1 | 500K | 0 | 0.276 | 0.344 | 0.252 | 0.258 | 0.237 | 0.263 | 0.252 | 0.260 | 0.254 | 0.261 |
| Model 1 | 500K | 0.5 | 0.359 | 0.486 | 0.256 | 0.261 | 0.629 | 0.694 | 0.259 | 0.269 | 0.543 | 0.563 |
| Model 2 | 50K | 0 | 0.275 | 0.399 | 0.252 | 0.263 | 0.232 | 0.294 | 0.244 | 0.270 | 0.251 | 0.270 |
| Model 2 | 50K | 0.5 | 0.250 | 0.442 | 0.304 | 0.325 | 0.329 | 0.419 | 0.244 | 0.274 | 0.329 | 0.346 |
| Model 2 | 500K | 0 | 0.270 | 0.303 | 0.253 | 0.255 | 0.250 | 0.260 | 0.256 | 0.257 | 0.257 | 0.258 |
| Model 2 | 500K | 0.5 | 0.313 | 0.346 | 0.305 | 0.308 | 0.339 | 0.354 | 0.256 | 0.259 | 0.332 | 0.335 |
| Model 3 | 50K | 0 | 0.270 | 0.441 | 0.432 | 0.455 | 0.225 | 0.314 | 0.259 | 0.277 | 0.261 | 0.277 |
| Model 3 | 50K | 0.5 | 0.293 | 0.445 | 0.447 | 0.467 | 0.410 | 0.501 | 0.255 | 0.283 | 0.268 | 0.289 |
| Model 3 | 500K | 0 | 0.257 | 0.307 | 0.254 | 0.255 | 0.247 | 0.262 | 0.258 | 0.259 | 0.258 | 0.259 |
| Model 3 | 500K | 0.5 | 0.275 | 0.321 | 0.442 | 0.443 | 0.444 | 0.457 | 0.269 | 0.270 | 0.279 | 0.280 |
| Model 4 | 50K | 0 | 0.303 | 0.479 | 0.548 | 0.582 | 0.522 | 0.690 | 0.527 | 0.614 | 0.560 | 0.632 |
| Model 4 | 50K | 0.5 | 0.251 | 0.581 | 0.338 | 0.369 | 0.757 | 1.030 | 0.312 | 0.384 | 0.681 | 0.788 |
| Model 4 | 500K | 0 | 0.294 | 0.321 | 0.548 | 0.555 | 0.538 | 0.571 | 0.548 | 0.560 | 0.556 | 0.566 |
| Model 4 | 500K | 0.5 | 0.319 | 0.384 | 0.332 | 0.334 | 0.722 | 0.768 | 0.328 | 0.337 | 0.684 | 0.700 |

The comparisons show that GAMI-NET and GAMI-Tree are comparable except when the FAST interaction filtering misses some interactions. Both are better than EBM. Xgboost is better only in the three-way interaction case since the other models cannot capture the higher-order term.

Next, we will compare the interpretation results among the GA2M models. We start from main effects. There are 10 true main effect variables in the model, $x_1 - x_{10}$. All algorithms capture these 10 variables as the 10 most important main effects. For the other redundant or irrelevant variables, GAMI-Tree and GAMI-Net do the best job in assigning low importance scores to those variables. There are two reasons:

1. First, in the round-robin training method used in EBM, all variables will be used regardless of whether they are truly important or not. However, GAMI-Tree selects only the best variable to model in each iteration, and it stops if model performance stops improving. This means the non-model variables will only be used few times in GAMI-Tree. In GAMI-Net, a pruning step is implemented, which keeps only the top $k-$most important terms. Therefore, most non-model variables have exactly zero importance.
2. Second, when the variables have correlation, the main-effect stage is more prone to assign importance to correlated, non-model variables. However, the iterative training in GAMI-Tree can reverse the false main effects captured in the first round, leading to close-to-zero importance for such redundant variables. GAMI-Net has a fine-tune stage where all main-effects and interactions are retrained simultaneously. This has the same effect as iterative training employed in GAMI-Tree.

To demonstrate the first point, consider Model 4 with 50K and $\rho = 0$. Since correlation is zero, all variables except $x_1$-$x_{10}$ are irrelevant and should receive close to zero importance score. However, from Figure 2, we can see EBM assigns relatively higher importance to those irrelevant variables. This is confirmed by the plot of the main effects for the top irrelevant variables in Figure 3. We can see EBM has a larger range than all other methods. Similar behavior has been observed for other models as well.

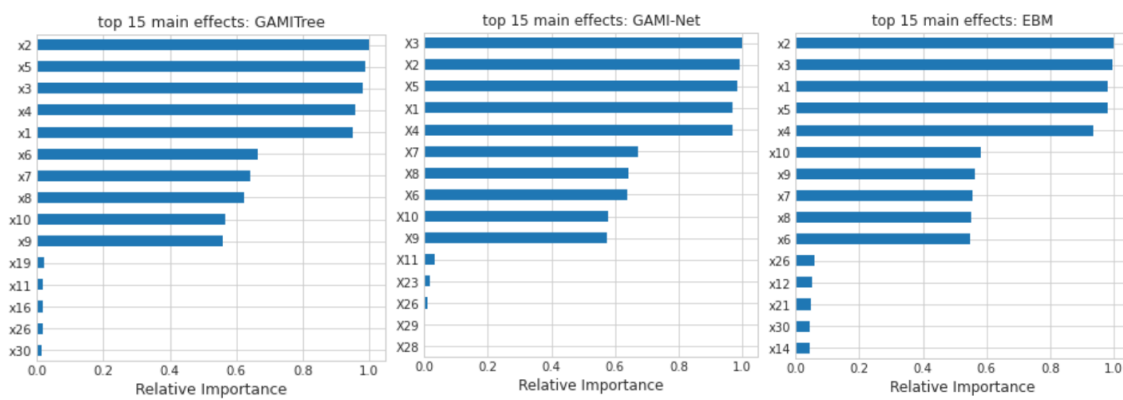

Figure 2. Main-effect importance for Model 4, 50K and rho=0

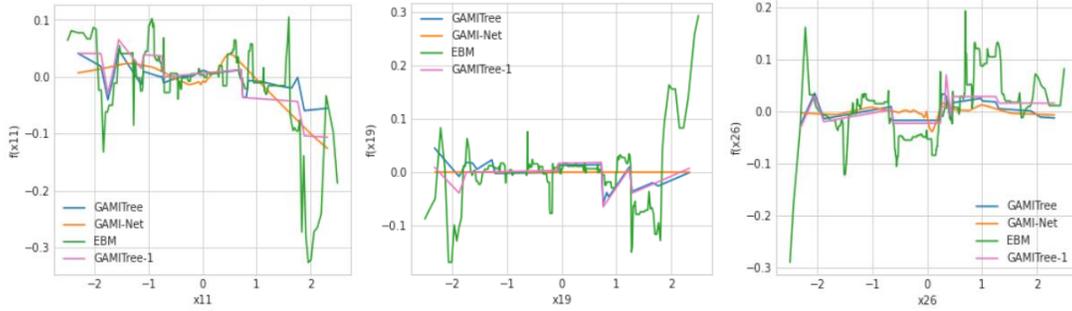

*Figure 3. Main effect plot for non-model variables, Model 4, 50K and rho=0*

To show the second point, consider again Model 4 with 50K but now $\rho = 0.5$. EBM assigns non-negligible importance to redundant variables ($x_{11}$-$x_{20}$), as we can see from Figure 4. Similar result is observed for GAMI-Tree-1 (not shown here). However, by iterating, GAMI-Tree effectively reduces the importance of these redundant variables to close to zero. This is supported by the plot of the main effects of $x_{18}$, as shown in the left plot in Figure 5. We can see the fake quadratic effect captured by EBM and GAMI-Tree-1. However, if we look at the second round in GAMI-Tree, referred to as GAMI-Tree-2, we find that the main effects of $x_{18}$ (Figure 5, right plot) is the opposite of the main effect captured in the first round. When adding them together, it eliminates the fake main effect of $x_{18}$. Similar behavior has been observed for $x_{19}$. For GAMI-Net, the fine-tune stage assigns close to zero importance to these variables.

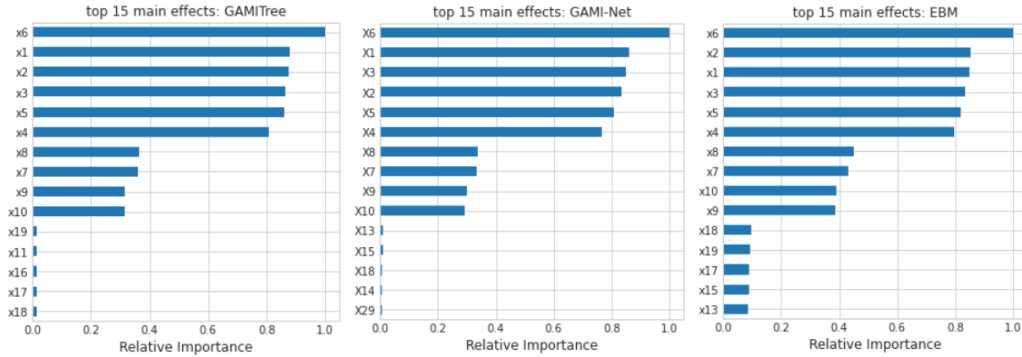

*Figure 4. Main-effect importance for Model 4, 50K and rho=0.5*

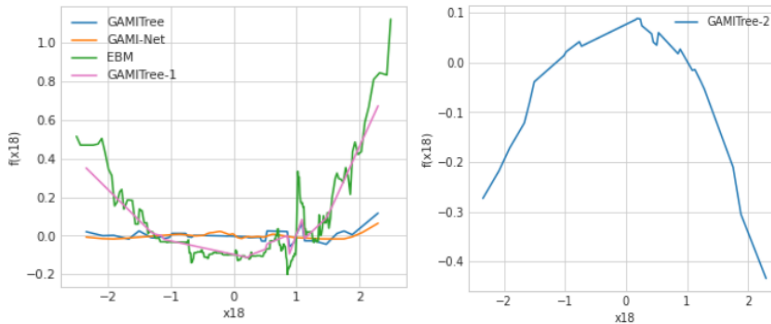

*Figure 5. Main-effect plot for non-model variables, Model 4, 50K and rho=0.5*

For the true model variables, the main effects from GAMI-Tree, GAMI-Net and EBM are very close for $\rho = 0$ case, except EBM is wigglier due to its piecewise constant nature and GAMI-Net is smooth. For $\rho = 0.5$ case, the iterative training in GAMI-Tree and fine-tune stage in GAMI-Net lead to more accurate results. Again, consider Model 4 with 50K and $\rho = 0.5$ scenario and focus on $x_9$ and $x_{10}$. In this case, variables $x_9$ and $x_{10}$ are purely additive since interactions only exist between $x_1 - x_6$. So, the true main effect is the function $x_j(x_j > 0), j = 9, 10$. Figure 6 shows the main effects from GAMI-Tree, GAMI-Net, EBM and GAMI-Tree-1. We can see both EBM and GAMI-Tree-1 show some uptick pattern in the negative region, whereas GAMI-Tree and GAMI-Net is close to the true form which is flat in the negative region. Similar pattern is observed for the larger 500K data. Therefore, GAMI-Net and GAMI-Tree give more accurate characterization of the main effects, with no distortion for the true model variables and negligible effect for the non-model variables (as we see previously).

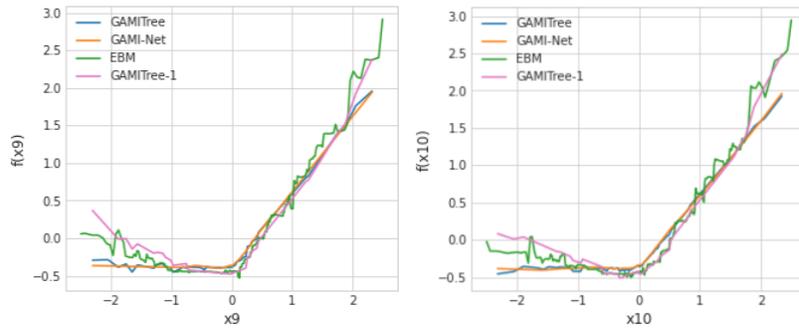

Figure 6. Main effect for x9 and x10, Model 4, rho=0.5, 50K (upper) and 500K (bottom)

Now consider model interpretation related with two-way interactions. First, we examine if each method captured all the true interaction pairs.

- For Models 1 and 4, all true interaction pairs are captured as the top ones by all models.
- For Model 2, $\rho = 0$, all eight true interaction pairs are captured as the top eight. However, for $\rho = 0.5$, EBM and GAMI-Net miss two true interaction pairs in their top 10 list: $0.25x_1x_2$ and $clip(x_7 + x_8, -1, 0)$, for both 50K and 500K sample sizes. For example, see Figure 7 for 50K case. This is due to the correlation among variables. It causes the "pure interaction" effect (after removing main-effects) for these two less-important interactions to be weaker and harder to identify, and some "surrogate interactions" (interactions that are not in the true model but mimicking the true interaction pairs due to variable correlation) for the six strong ones could rank higher during interaction filtering. For GAMI-Tree, its first round (GAMI-Tree-1) also suffers from this surrogate interaction issue and misses the above two interactions. However, its second-round redoes the interaction filtering. Because interactions for the first six pairs have already been accounted for in the first round, the surrogate interaction pairs are no longer significant, and it is easy to identify the missed interactions. As a result, the second round accurately picked up the two missed interaction as the two most important interaction pairs, and GAMI-Tree was able to capture all eight true interaction pairs and list them as top eight correctly.
- For Model 3, $\rho = 0.5$, EBM and GAMI-Net both miss the two sine function related interactions, $x_5$-$x_6$ and $x_7$-$x_8$, whereas GAMI-Tree captures all four true interactions. For example, see Figure 8 for 50K data case. This is because the 4-quandrant model used in FAST algorithm cannot capture the highly nonlinear sine function well, so it misses out on these two interactions. On the other hand, the interaction tree we used in GAMI-Tree interaction filtering can capture this well.

- For Model 3, 50K, $\rho = 0$, GAMI-Net misses the two sine function interaction due to the limitation of FAST algorithm mentioned earlier, resulting in a worse model performance.
- Finally, the true two-way interaction effects captured by all methods are similar. Figure 9 shows the two-way interactions from GAMI-Tree (top), GAMI-Net (second), EBM (third) and the true model after orthogonalization (bottom). We plot $f(x_j, x_k)$ when fixing $x_k$ at a set of quantile values. Since EBM and GAMI-Net only capture six true interactions, only those six are shown. We can see they have similar patterns in general, with GAMI-Net being smooth and others showing some wiggly patterns.

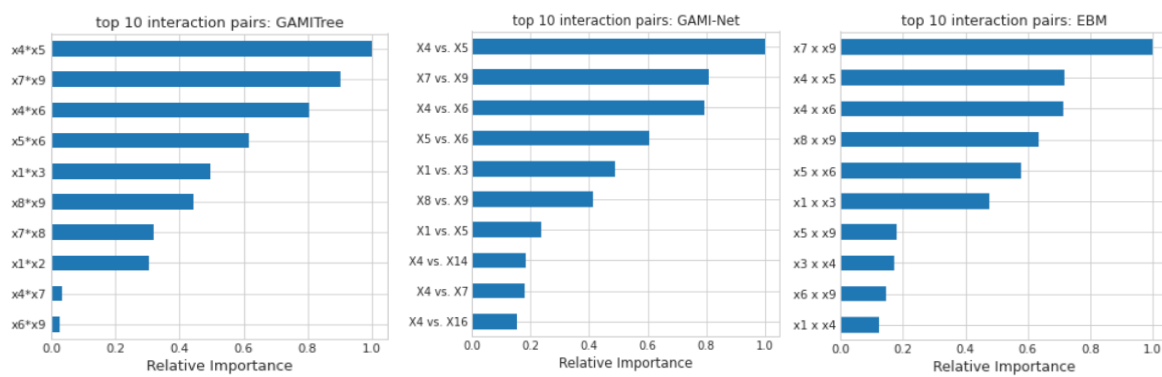

*Figure 7. Interaction importance for Model 2, 50K and rho=0.5*

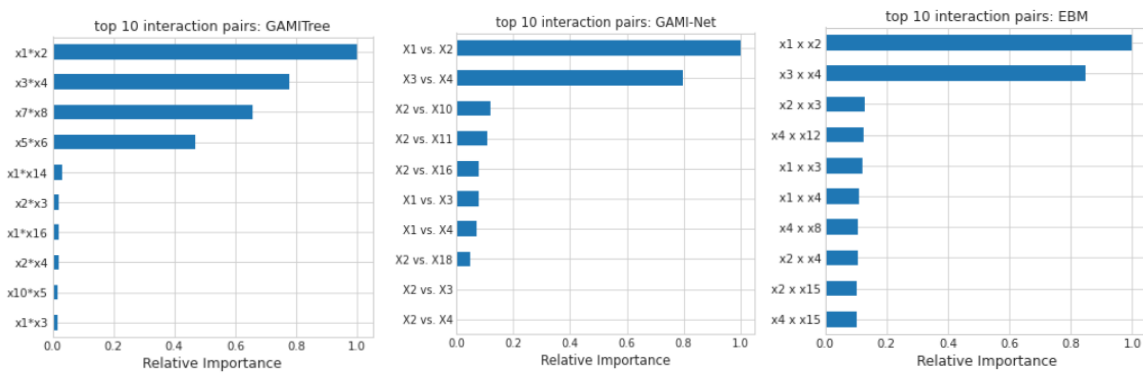

*Figure 8. Interaction importance for Model 3, 50K and rho=0.5*

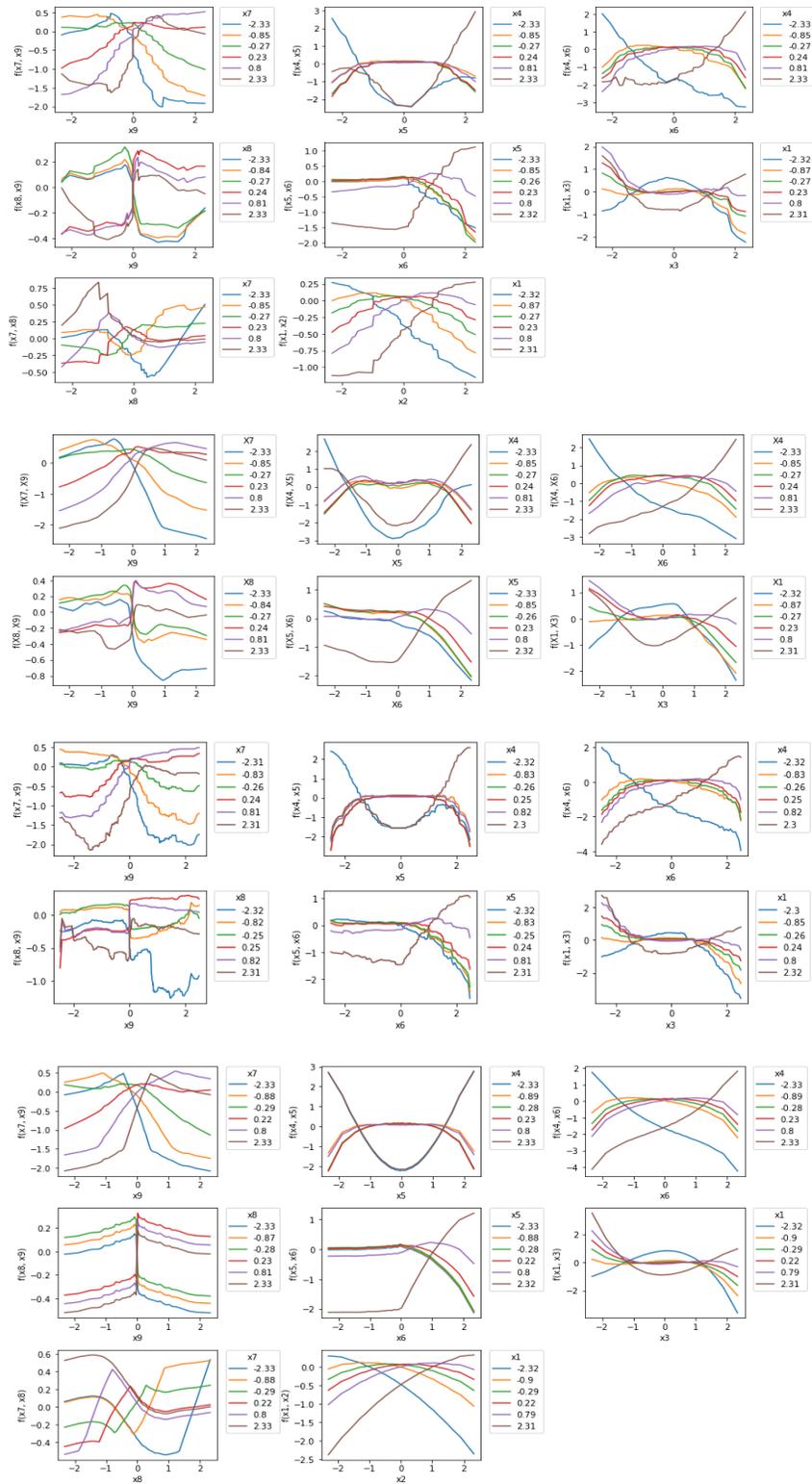

*Figure 9. Interaction effect plot for GAMI-Tree (top), GAMI-NET (second), EBM (third) and true model (bottom), for Model 2, 50K and rho=0.5*

## 5.2.2 Binary Case

The results for the binary case were qualitatively similar to the continuous case, but noisier and less significant. For example, the AUC performances among all models are very close and the improvement from GAMI-Tree-1 to GAMI-Tree is tiny. The main-effects and interactions captured are noisier due to smaller signal to noise ratio in binary response. Details can be found in Appendix B section.

# 6. Applications to Real Data

## 6.1 Bike sharing

This is a public data hosted on UCI machine learing repository. It has around 17,000 hourly bike rental counts from 2011 to 2012, with corresponding time (by hour), weather and season information. The goal is to predict hourly bike rental counts. We used log counts as response and the following 11 variables as predictors: *yr* (year, 1 if 2012 and 0 if 2011); *mnth* (month = 1 to 12); *hr* (hour = 0 to 23); *holiday* (1 if yes and 0 otherwise); *weekday* (0 = sunday to 6 = saturday); *workingday* (1 if working and 0 if weekend or holiday); *season* (1:winter; 2:spring, 3:summer, 4:fall); *weathersit* (1:clear, 2: misty+cloudy; 3: light snow; 4 :heavy rain); *temp* (normalized to be within 0 and 1); *hum* (humidity) and *windspeed*. There are some identifiability issues here as *working day* is completely determined by *holiday* and *weekday*.

The data was split into 50% training, 25% validation and 25% testing, and the following algorithms were fit: xgboost, GAMI-Net, GAMI-Tree and EBM. We used the same tuning/training settings as in Section 5. The training and testing mse for all models are listed in Table 7. xgboost is the best, GAMI-Tree is second, followed by EBM and GAMI-Net. There are also some improvements from GAMI-Tree-1 to GAMI-Tree.

*Table 7. Train and test mse for bike sharing data*

|  | xgboost | GAMI-Net | GAMI-Tree | GAMITree-1 | EBM |
|---|---|---|---|---|---|
| **train_mse** | 0.055 | 0.132 | 0.108 | 0.121 | 0.116 |
| **test_mse** | 0.099 | 0.119 | 0.103 | 0.111 | 0.107 |

Figure 10 shows the importance ranking for the 11 main effects. All algorithms yield similar rankings with some slight change of orders. For GAMI-NET, the bottom three variables have exactly zero importance. This is due to the pruning step we mentioned.

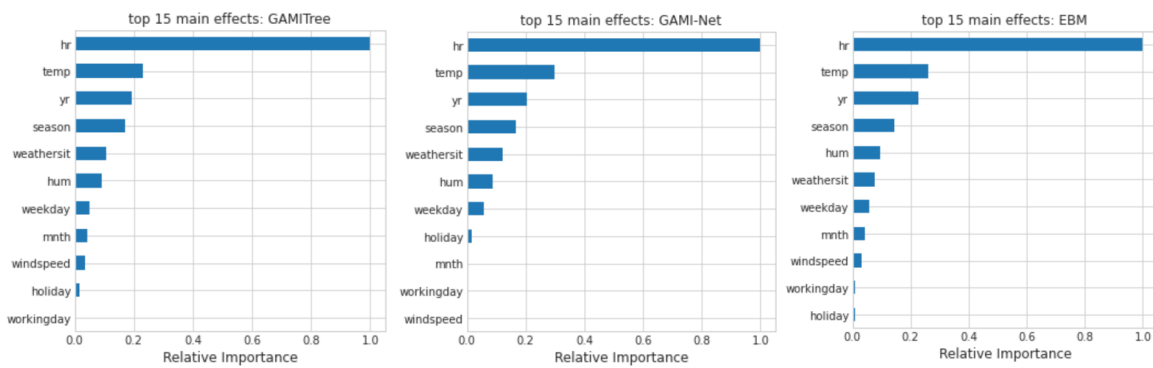

*Figure 10. Main-effect importance for bike sharing data*

Figure 11 shows the main effect plot for EBM, GAMI-Net and GAMI-Tree. They overlap well, particularly EBM and GAMI-Tree. The biggest difference seems to come from weathersit variable, at value 4 = heavy rain. However, only 3 out of the total of 17379 records have this values, so this is unreliable. GAMI-Net does not show the double peaks for the *hour* variable, but it shows double peaks in the interaction effect in Figure 13. So this is due to how main-effects and interactions are decomposed in GAMI-Net. Finally, the main-effect plots of *mnth*, *windspeed* and *workingday* are flat for GAMI-NET, which is consistent with their importance scores.

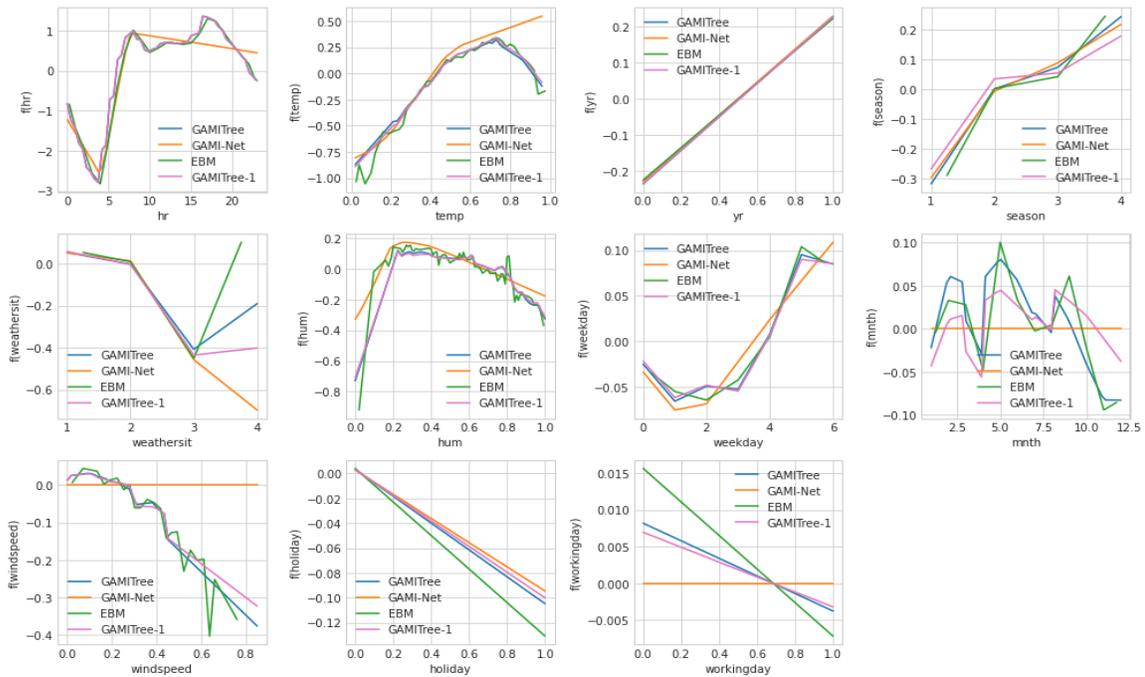

*Figure 11. Main effect plot for bike sharing data*

The top ten interactions from GAMI-Tree, GAMI-Net, and EBM are shown in Figure 12. The top two pairs identified by all three are the same. In fact, GAMI-Tree and EBM have the same top three. There are differences for the weaker interactions. For example, GAMI-Net does not have *yr-mnth* interaction. EBM ranks *hr-temp* interaction as 4[th] while GAMI-Tree ranks it as 7[th]. On the other hand, GAMI-Tree ranks *hr-hum* as the 4[th] but EBM ranks it as 6[th].

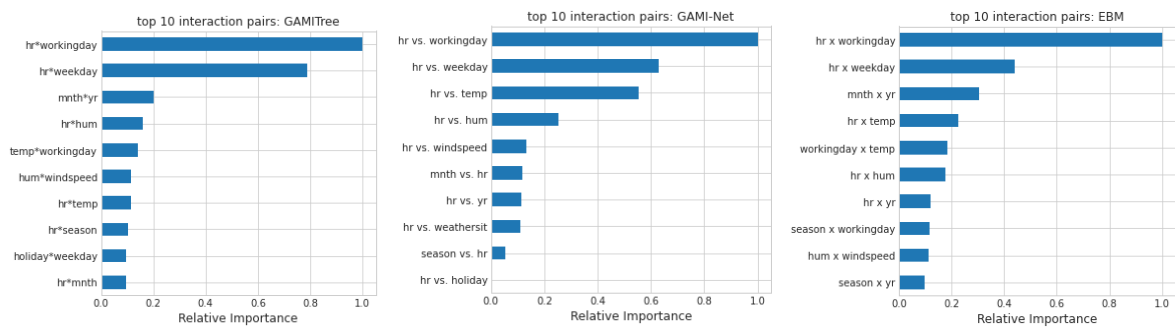

*Figure 12. Interaction importance for bike sharing data*

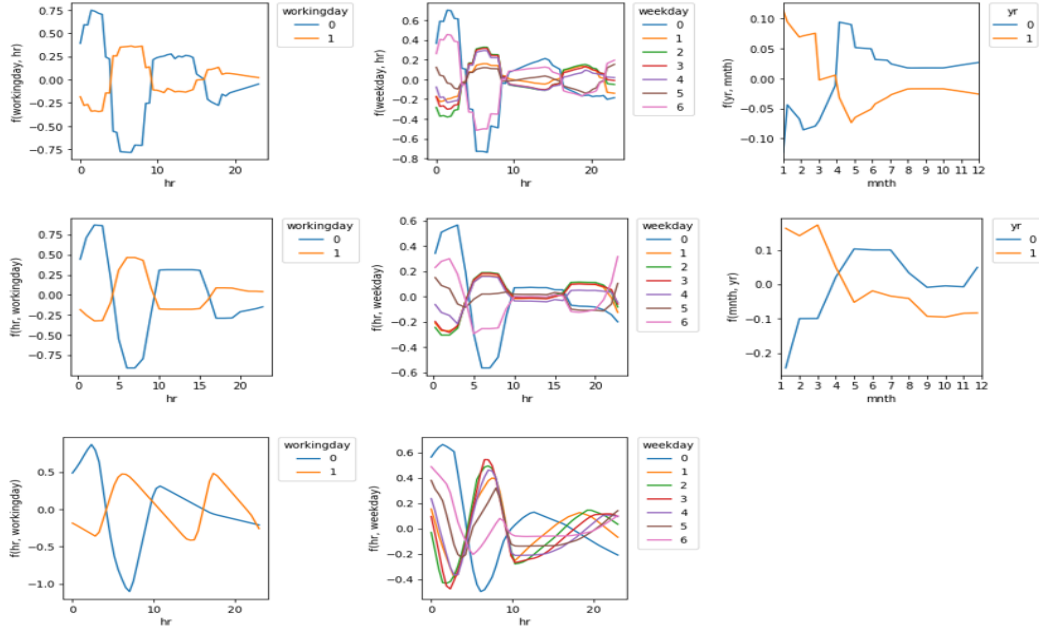

*Figure 13. Selected interaction plots for bike sharing data, GAMI-Tree (top), EBM (middle) and GAMI-Net (bottom)*

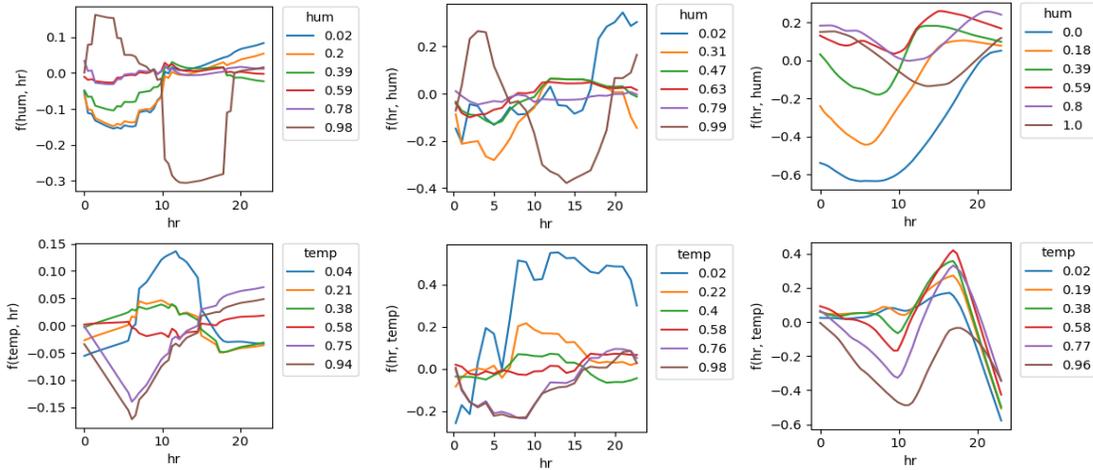

*Figure 14. Selected interaction plots for bike sharing data, GAMI-Tree (left), EBM (middle) and GAMI-Net (right)*

Figure 13 shows the top 3 interactions from GAMI-Tree, EBM and the top 2 interactions from GAMI-Net. The patterns among EBM and GAMI-Tree look very similar. Since workingday = 1 is highly correlated with weekday being 1 to 5 (except some holidays), the first two interaction pairs are very similar. They both indicate that, on non-working days, bike rentals peak between 10 am to 16 pm, whereas for working days, bike rentals peak in the morning and afternoon during rush hours. There are some changes in the monthly patterns for the two different years, but the effect is much weaker compared to the first two interaction pairs. GAMI-Net shows similar pattern for *hr* and *workingday/weekday* interaction, except the afternoon peak for *hour* on working days is more obvious. This is related with what we saw in Figure 11, where the main effect plot for *hour* misses the afternoon peak. So the three methods can have some differences due to how main-effects and interactions are decomposed.

Figure 14 shows the intearction of *hr-hum* and *hr-temp*, the 4[th] pair from GAMI-Tree and EBM, respectively, as well as the third and fourth pair from GAMI-Net. The interaction for *hr-hum* is similar among GAMI-Tree and EBM. Most of this interaction is related with high humidity, where it reduces bike rental between 10am-18pm and increases it after midnight[3] (compared to the 'average behavior' captured in the main-effects). However, GAMI-Net shows a quite different pattern. The interaction for *hr-temp* is weak for GAMI-Tree and EBM, but quite strong for GAMI-Net. Aside from when temp is really low (below 0.05, which accounts for only less than 0.2% of data), EBM and GAMI-Tree have similar patterns. They both show bike rentals increase when temperature is moderate: 8 am-12pm when it is cool and 6pm to midnight when it is hot. GAMI-Net assigns a high importance for this pair of interaction, and the pattern does not agree well with GAMI-Tree or EBM. There are two possible reasons for theses differences: correlation impact, or how the effects are decomposed into main effect and interactions.

## 6.2 Home lending

This applications deals with residential mortgage accounts. The response variable is a "troubled" loan indicator: 1 if the loan is in trouble state and 0 otherwise. The term trouble is defined as any of the following events: bankruptcy, short sale, 180 or more days of delinquency in payments, etc. The goal is to predict if a loan will be in trouble at a future prediction time based on account information from the current time (called snapshot time) and macro-economic information at the prediction time. The time interval between prediction time and current time is called prediction horizon (see Figure 15 for an illustration).

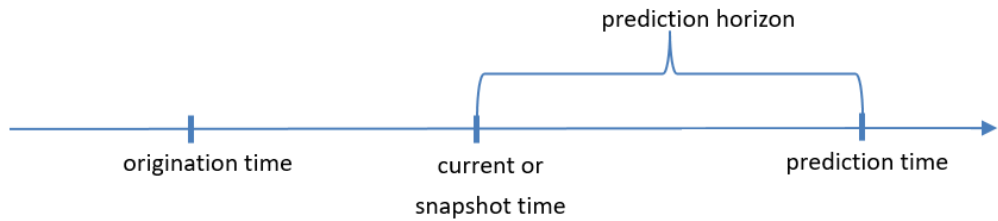

*Figure 15. Loan origination, current (snapshot) and prediction times*

There are over 50 predictors, including macro-economic variables (unemployment rate, house price index, and so on), static loan characteristic variables at the origination time (fixed 15/30 year loan, arm loan, balloon loan, etc), and dynamic loan characteristic variales (snapshot fico, snapshot delinquency status, forecasted loan-to-value ratio, etc). For model interpretation purpose, we removed some variables which are highly correlated, and used 44 of them to fit our models. The important variables are listed in Table 8.

*Table 8. Variable definitions for home lending data*

| Variable | Definition |
|---|---|
| horizon | prediction horizon (difference between prediction time and snapshot time) in quarters |
| snap_fico | credit score (FICO score) at snapshot time |
| orig_fico | credit score (FICO score) at loan origination |
| snap_ltv | loan to value (ltv) ratio at snapshot time |
| fcast_ltv | loan to value (ltv) ratio forecasted at prediction time |
| orig_ltv | loan to value ratio at origination |

---

[3] Note interactions captured here does not include main-effects, so they should be interpreted as an adjustment to the average behavior captured by the main-effects.

| orig_cltv | combined ltv at origination |
|---|---|
| snap_early_delq_ind | early delinquency (no min payments for a few months) indicator: 1 means loan has early delinquency status at snapshot time; 0 means loan is current or has late delinquency status. 7.7% observations are early delinquent. |
| snap_late_delq_ind | late delinquency indicator (loan is delinquent for longer time, close to default) indicator: 1 means loan has late delinquency status at snapshot time; 0 means loan is current or has early delinquency status. Only 0.2% observations are late delinquent. |
| pred_loan_age | age of loan (in months) at prediction time |
| snap_gross_bal | gross loan balance at snapshot time |
| orig_loan_amt | total loan amount at origination time |
| pred_spread | spread (difference between note rate and market mortgage rate) at prediction time |
| orig_spread | spread at origination time |
| orig_arm_ind | Indicator: 1 if loan is adjustable-rate mortgage (ARM); 0 otherwise |
| pred_mod_ind | modification indicator: 1 means prediction time before 2007Q2 (financial crisis); 0 if after |
| pred_unemp_rate | unemployment rate at prediction time |
| pred_hpi | house price index (hpi) at prediction time |
| orig_hpi | hpi at origination time |
| pred_home_sales | home sales data at prediction time |
| pred_rgdp | real GDP at prediction time |
| pred_totpersinc_yy | total personal income growth (from year before prediction to prediction time) |

*Table 9. Train and test auc, logloss for home lending data*

|  | xgboost | GAMITree | GAMI-Net | GAMITree-1 | EBM |
|---:|---:|---:|---:|---:|---:|
| **train_auc** | 0.906 | 0.869 | 0.849 | 0.861 | 0.865 |
| **train_logloss** | 0.0415 | 0.0451 | 0.0467 | 0.0460 | 0.0455 |
| **test_auc** | 0.857 | 0.858 | 0.851 | 0.855 | 0.855 |
| **Test_logloss** | 0.0451 | 0.0451 | 0.0457 | 0.0455 | 0.0454 |

We selected a subset of 1 million observations from the original dataset for one of the portfolio segments. The data was split into 50% training, 25% validation and 25% testing. Again, we fitted all four algorithms: xgboost, GAMI-Net, GAMI-Tree, and EBM. We used the same tuning/training settings as in Section 5. The training and testing auc for all models are listed in Table 9. The performance of xgboost, GAMI-Tree and EBM are all comparable, with GAMI-Tree being the best. GAMI-Net is slightly worse. There are slight improvements from GAMI-Tree-1 to GAMI-Tree.

Figure 16 shows the importance ranking for the top 10 main effects. The rankings among all models are close with some small differences. For example, GAMI-Tree ranks horizon as 7[th] important main effect, whereas GAMI-Net ranks it as 10[th] and EBM does not rank it as one of the top 10 main effects; on the other hand, EBM/GAMI-Net ranks interest only indicator as the 9[th] important main effect, whereas GAMI-Tree does not rank it as one of the top 10 main effects. Comparing GAMI-Net and EBM, they are very consistent except the 10[th] variable is different and some slight change in the ranking.

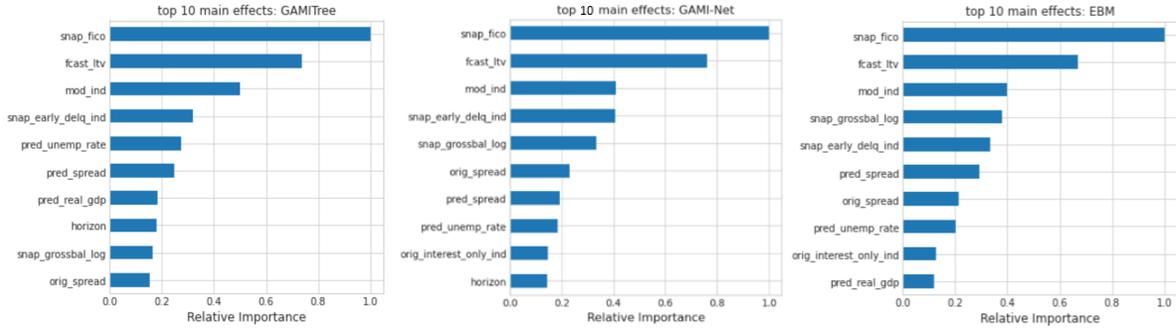

*Figure 16. Main-effect importance for home lending data*

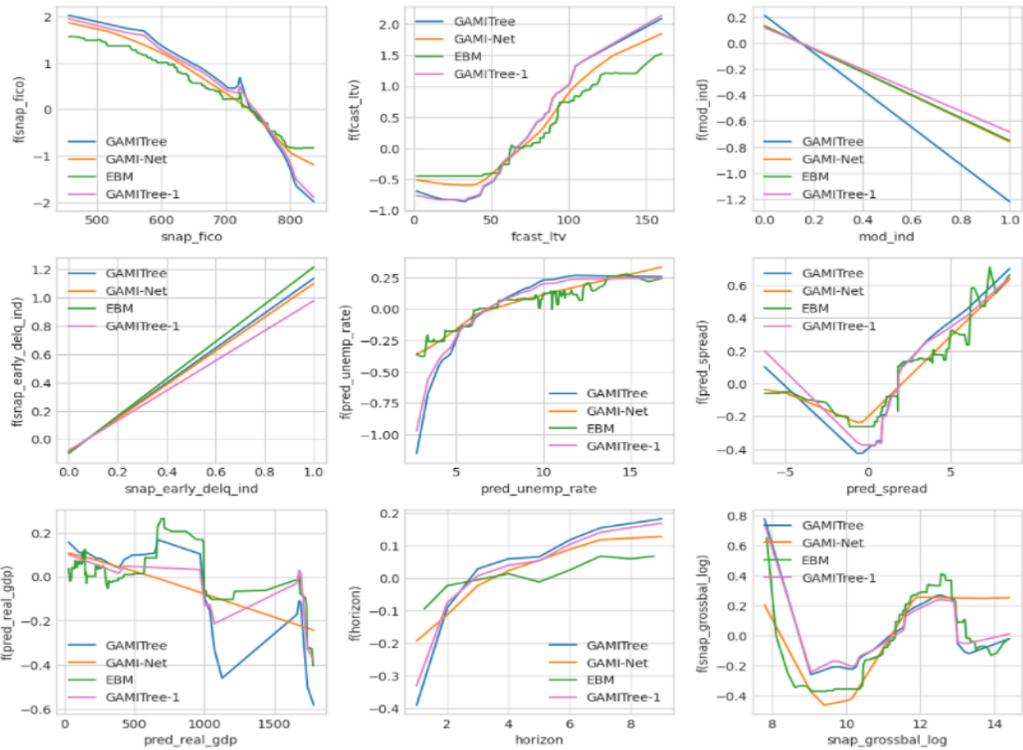

*Figure 17. Main effect plot for home lending data*

Figure 17 shows the main effect plot for all models for the top 9 variables in GAMI-Tree. GAMI-Tree and GAMI-Tree1 are very close and both show stronger main effects for a few variables, including snapshot fico, forecasted ltv, unemployment rate and horizon. Some of the differences can be explained by the purification step we use in GAMI-Tree, which we will discuss in more details later.

The top 10 interactions from GAMI-Tree, GAMI-Net and EBM are shown in Figure 18. The top four interactions from GAMI-Tree are: mod_ind and fico, ltv and fico, unemployment rate and fico, horizon and early delinquency indicator. Those interactions make sense from the subject-matter perspective and have been seen in other studies. EBM did not have the unemployment rate vs fico interaction and mod_ind vs fico interaction. On the other hand, EBM captures multiple pairs of interactions related with late delinquency indicator, most importantly, the interaction among horizon and late delinquency indicator. While this variable pair indeed has interaction, late delinquency is a very rare event (only 0.2% observations in total), and GAMI-Tree does not rank it as top 10. For GAMI-Net, the top 10 interactions filtered by FAST algorithm again has a lot of late delinquency indicator related interactions, but the fine tune step pruned 4 of them, keeping a total of 6 interactions. The top 2 interactions are ltv and fico, horizon and early delinquency indicator, which are high ranking interactions in all algorithms; however, it does not have unemployment rate and fico, or spread and fico interactions. Increasing the number of interactions in filtering step allows it to capture those interaction pairs, and have better model performance.

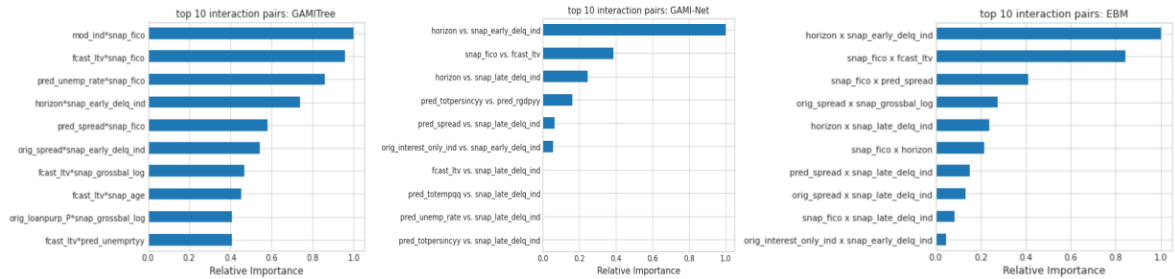

*Figure 18. Interaction importance for home lending data*

Figure 19 shows the top three of the common interaction pairs from GAMI-Tree and EBM, two of which are also top two in GAMI-Net. The patterns look very similar with some differences. For example, for the interaction between horizon and early delinquency indicator, GAMI-Tree shows that the effect of horizon is almost flat when the loan is not in early delinquency state, whereas EBM and GAMI-Net show an increasing trend. Recall that in Figure 17, the main-effect of horizon is flatter for EBM and GAMI-Net compared to GAMI-Tree, we can see that the difference is due to how the main-effect and interactions are decomposed in each model. Particularly, the interaction effect from EBM and GAMI-Net still has some main effect on the horizon variable, this results in flatter trend for horizon in the main-effect plot. GAMI-Tree uses a post-hoc orthogonalization step to make sure interactions do not contain any main-effects, whereas EBM and GAMI-Net (uses a clarity penalty) does not guarantee this.

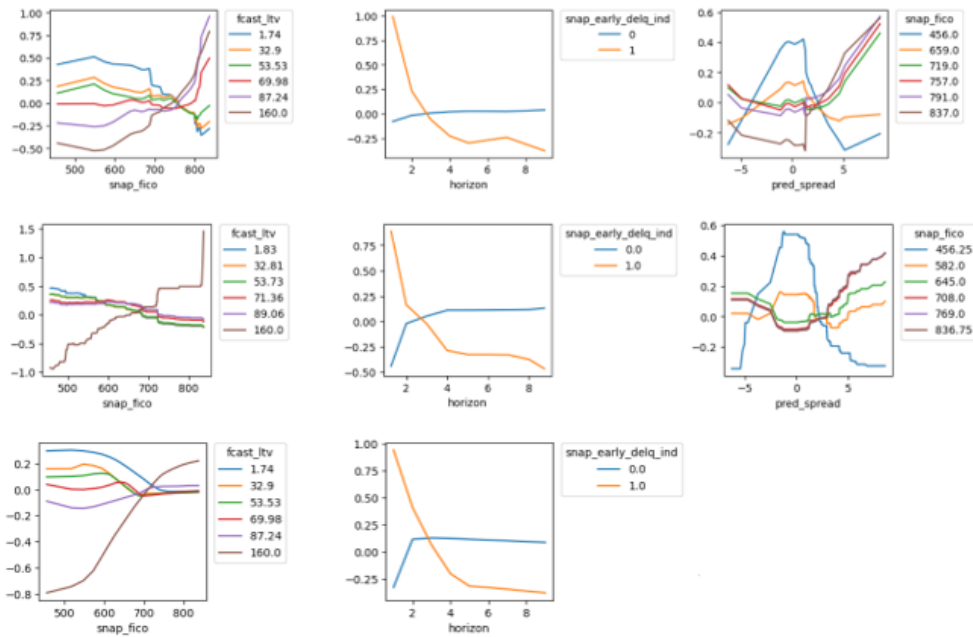

*Figure 19. Interaction plot for home lending data, GAMI-Tree (top), EBM (middle), GAMI-Net (bottom)*

To further demonstrate the difference orthgonalization has made, Figure 20 shows the main effects importance for GAMI-Tree with and without orthogonlization. We can see the top 10 variables of unpurified GAMI-Tree are same as EBM (with slight change in ranking), and neither contains horizon variable. In addition, the main-effect plots are more similar among EBM and unpurified GAMI-Tree, as shown in Figure 21. In particular, the main-effect of horizon becomes small for unpurified GAMI-Tree. This indicates that the main-effect of horizon we see from (purified) GAMI-Tree comes from orthogonalization of interaction effects, and it also explains why the effect of horizon is very flat for early delinquency = 0 accounts in Figure 19.

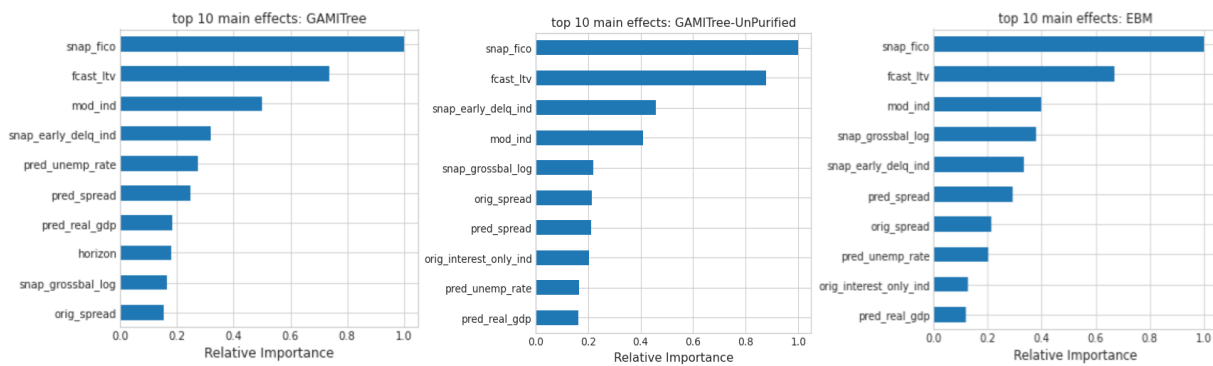

*Figure 20. Main-effect importance for purified GAMI-Tree (left), Unpurified GAMI-Tree (middle) and EBM (right)*

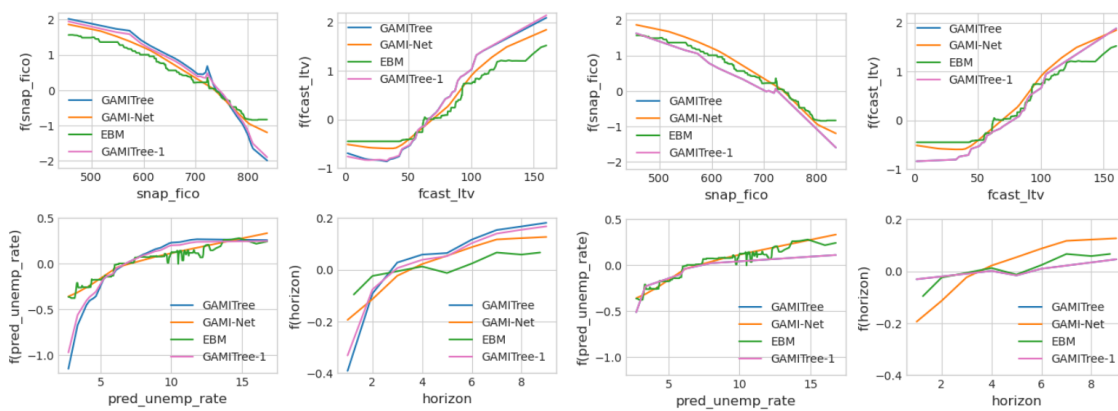

*Figure 21. Main-effect plots for purified GAMI-Tree (left) and unpurified GAMI-Tree (right)*

## 7. Summary

Interpretability is a very important attribute of model performance, and it allows stake holders to understand and manage model risks. This paper contributes to the literature on inherently-interpretable models using effective methodology and fast algorithms to estimate main-effects and two-way interactions nonparametrically. GAMI-Tree performs comparably or better than EBM in terms of predictive performance and is able to identify the interactions more accurately. This is due to several novel features: improved base learners for estimating non-linear main effects and interactions, new interaction filtering method which captures interactions more accurately, a new iterative training method which converges to more accurate models and an orthogonalization method to make sure interactions and main effects are hierarchically orthogonal. We have used simulation studies and real applications to demonstrate the usefulness of GAMI-Tree in terms of model performance and model interpretation.

# Acknowledgements

We thank Zebin Yang for his generous help in running GAMI-Net algorithm for the simulated and real data cases.

# Appendix A: Proof for Hierarchical Orthogonality

**Theorem**: Given a function $f(x)$, $x = (x_1, \ldots x_p)$, if $f_{-1}(x_{-1}), \ldots, f_{-p}(x_{-p})$ are solutions that minimize $\int \left( f(x) - \sum_{j=1}^{p} f_{-j}(x_{-j}) \right)^2 w(x) dx$, then the residual $\tilde{f}(x) = f(x) - \sum_{j=1}^{p} f_{-j}(x_{-j})$ is orthogonal with any lower dimensional function $h(x_{-j})$, ie, $\int \tilde{f}(x) h(x_{-j}) w(x) dx = 0$ for any $h(x_{-j})$ and any $1 \leq j \leq p$.

**Proof**: Suppose there exists some $h(x_{-j})$ such that $b = \int \tilde{f}(x) h(x_{-j}) w(x) dx > 0$, then consider the new solution to the least weighted sum of squared problem, $\int \left( f(x) - \left( \sum_{j=1}^{p} f_{-j}(x_{-j}) + \lambda h(x_{-j}) \right) \right)^2 w(x) dx = \int \left( \tilde{f}(x) - \lambda h(x_{-j}) \right)^2 w(x) dx = \int \left( \tilde{f}(x) \right)^2 w(x) dx + \lambda^2 \int h^2(x_{-j}) w(x) dx - 2\lambda \int \tilde{f}(x) h(x_{-j}) w(x) dx = \int \left( \tilde{f}(x) \right)^2 w(x) dx + \lambda^2 \int h^2(x_{-j}) w(x) dx - 2\lambda b$. We can choose any small positive $\lambda < \frac{2b}{\int h^2(x_{-j}) w(x) dx}$, such that $\int \left( \tilde{f}(x) - \lambda h(x_{-j}) \right)^2 w(x) dx < \int \left( \tilde{f}(x) \right)^2 w(x) dx$, this leads to contradiction. Similarly, we can prove the case when $\int \tilde{f}(x) h(x_{-j}) w(x) dx < 0$.

# Appendix B: Results for Binary Simulation Cases

For binary case, the training and testing AUCs for the different models are listed in Table 10 below. We can see all models have close AUC, and GAMI-Tree has same or slightly better auc than all other algorithms in most cases. The difference between GAMI-Tree and GAMI-Tree-1 is tiny compared to continuous case. In terms of overfitting, GAMI-Net has smallest gaps between training and testing auc's from Table 10, followed by GAMI-Tree.

*Table 10. Train and test AUCs for binary simulation cases*

|  | N | rho | xgboost | | GAMI-Net | | GAMI-Tree | | GAMI-Tree-1 | | EBM | |
|---|---|---|---|---|---|---|---|---|---|---|---|---|
|  |  |  | train | test | train | test | train | test | train | test | train | test |
| **Model 1** | 50K | 0 | 0.948 | 0.889 | 0.895 | 0.888 | 0.906 | 0.890 | 0.905 | 0.890 | 0.924 | 0.892 |
| **Model 1** | 50K | 0.5 | 0.983 | 0.963 | 0.966 | 0.964 | 0.971 | 0.964 | 0.967 | 0.963 | 0.971 | 0.962 |
| **Model 1** | 500K | 0 | 0.923 | 0.894 | 0.897 | 0.893 | 0.903 | 0.897 | 0.902 | 0.896 | 0.909 | 0.897 |
| **Model 1** | 500K | 0.5 | 0.969 | 0.965 | 0.965 | 0.965 | 0.966 | 0.966 | 0.965 | 0.964 | 0.966 | 0.964 |
| **Model 2** | 50K | 0 | 0.948 | 0.913 | 0.920 | 0.918 | 0.927 | 0.918 | 0.925 | 0.918 | 0.932 | 0.916 |
| **Model 2** | 50K | 0.5 | 0.981 | 0.962 | 0.963 | 0.964 | 0.966 | 0.964 | 0.964 | 0.963 | 0.968 | 0.962 |
| **Model 2** | 500K | 0 | 0.927 | 0.917 | 0.920 | 0.918 | 0.921 | 0.919 | 0.921 | 0.919 | 0.923 | 0.919 |
| **Model 2** | 500K | 0.5 | 0.967 | 0.963 | 0.964 | 0.963 | 0.965 | 0.964 | 0.964 | 0.963 | 0.965 | 0.963 |
| **Model 3** | 50K | 0 | 0.938 | 0.897 | 0.904 | 0.902 | 0.913 | 0.903 | 0.911 | 0.903 | 0.916 | 0.898 |
| **Model 3** | 50K | 0.5 | 0.975 | 0.950 | 0.953 | 0.951 | 0.957 | 0.953 | 0.956 | 0.953 | 0.960 | 0.951 |

| Model 3 | 500K | 0   | 0.917 | 0.901 | 0.908 | 0.905 | 0.908 | 0.905 | 0.908 | 0.905 | 0.911 | 0.903 |
|---------|------|-----|-------|-------|-------|-------|-------|-------|-------|-------|-------|-------|
| Model 3 | 500K | 0.5 | 0.959 | 0.953 | 0.953 | 0.953 | 0.954 | 0.954 | 0.954 | 0.954 | 0.956 | 0.953 |
| Model 4 | 50K  | 0   | 0.962 | 0.909 | 0.920 | 0.916 | 0.927 | 0.915 | 0.922 | 0.913 | 0.931 | 0.909 |
| Model 4 | 50K  | 0.5 | 0.977 | 0.951 | 0.956 | 0.954 | 0.960 | 0.954 | 0.958 | 0.953 | 0.963 | 0.951 |
| Model 4 | 500K | 0   | 0.931 | 0.916 | 0.917 | 0.915 | 0.919 | 0.915 | 0.915 | 0.913 | 0.918 | 0.911 |
| Model 4 | 500K | 0.5 | 0.963 | 0.955 | 0.955 | 0.955 | 0.955 | 0.956 | 0.954 | 0.955 | 0.955 | 0.954 |

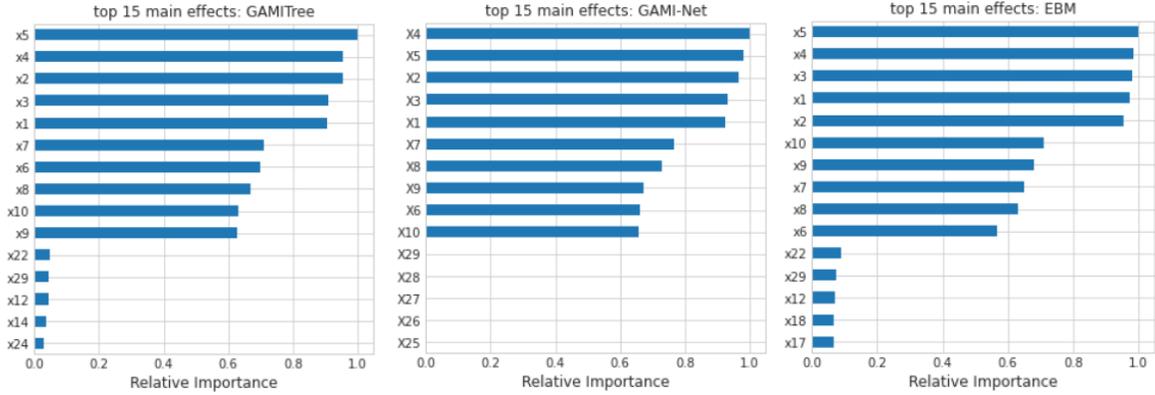

*Figure 22. Main effect importance for Model 4, 50K and rho=0*

For model interpretation, the conclusions are similar to those in the continuous case but less obvious in some cases. For main effects, we find GAMI-Tree and GAMI-Net give a more accurate description. First, they assign smaller importance to non-model variables. Figure 22 shows the main effect importance for model 4, 50K and $\rho = 0$. We can see GAMI-Net's pruning step keeps only the 10 true model variables, while dropping all non-model variables. GAMI-Tree still keeps the non-model variables, but it assigns smaller importance to them compared to EBM. Similar behavior has been observed for other models.

Second, as we mentioned in the continuous case, in the correlated cases, the iterative training in GAMI-Tree and fine-tune stage in GAMI-Net lead to a more accurate model than the simple two-stage algorithm. This is less obvious in the binary case than in the continuous case due to smaller signal-to-noise ratio (consistent with the less performance boost as we see in Table 10). However, for the larger 500K data, this effect can still be seen clearly. For example, Figure 23 shows the main effect importance for Model 4 with 500K and $\rho =0.5$. We can see the effect for those non-model variables (for example, $x_{13}$ and $x_{15}$) are smaller in GAMI-Tree/GAMI-Net compared to EBM and GAMI-Tree-1, and the effect of $x_9$ and $x_{10}$ are captured more accurately, as we see in Figure 24.

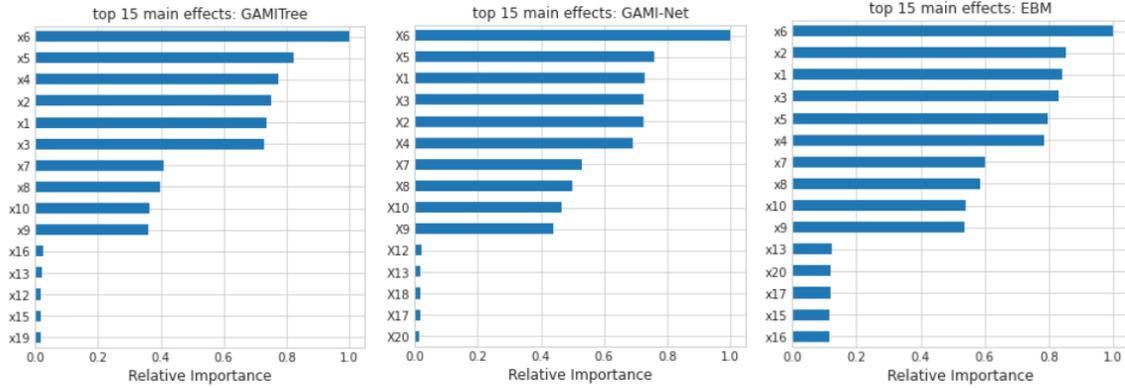

*Figure 23. Main effect importance for Model 4, 500K and rho=0.5*

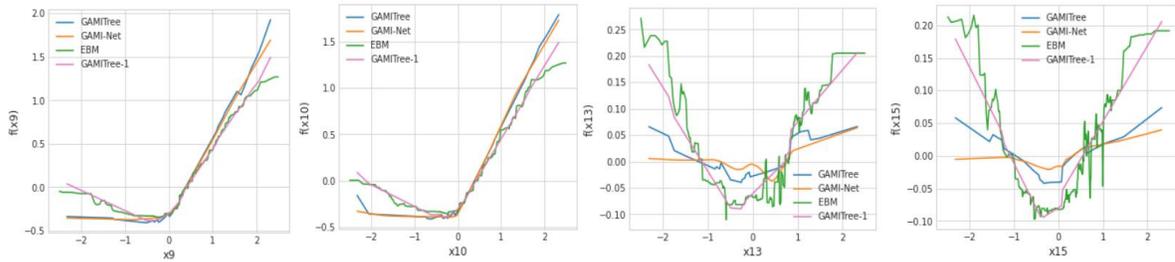

*Figure 24. Main effect for Model 4, 500K and rho=0.5*

Again, the conclusions for interpreting two-way interactions are similar to continuous case. We start with the question of whether the algorithms can capture all the true interaction pairs. For Model 1 and Model 4, all true interaction pairs are captured as the top ones. For Model 2, all methods do well for $\rho = 0$. However, for $\rho = 0.5$, we see similar behavior as in the continuous case. Specifically, for 50K sample size, EBM misses three true interaction pairs ($x_1$-$x_3$, $x_5$-$x_6$ and $x_7$-$x_8$) in its top 10 list. GAMI-Net misses three different pairs ($x_1$-$x_3$, $x_1$-$x_2$ and $x_7$-$x_8$), whereas GAMI-Tree captures all eight in its top eight list (see Figure 25). In fact, the first round in GAMI-Tree still misses one interaction pair ($x_1$-$x_3$), but the second round of interaction filtering helps pick up the missed pair. We see similar results for 500K: both EBM and GAMI-Net miss three pairs, and GAMI-Tree does not miss any. For Model 3, GAMI-Tree is able to capture all four true interactions, even in the first round. Our well-tuned EBM only misses the $x_3$-$x_4$ interaction for $\rho = 0.5$ case, but GAMI-Net (which uses default EBM for interaction filtering) misses additional interactions, particularly the sine related ones. For example, see Figure 26 for 50K data case.

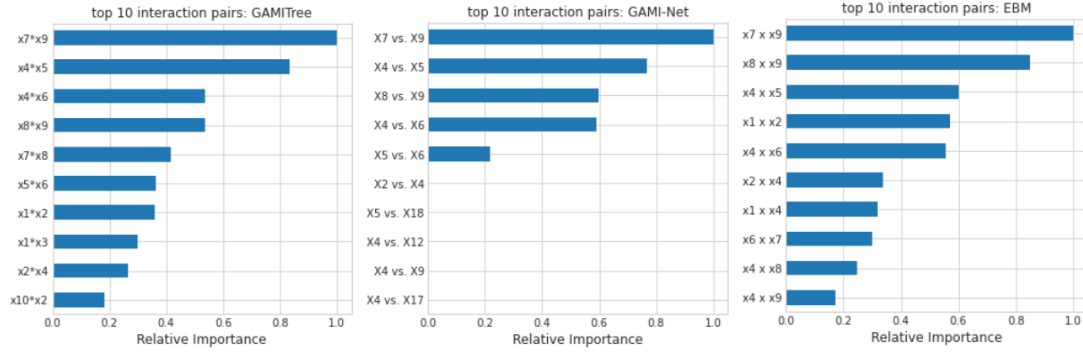

*Figure 25. Interaction importance for Model 2, 50K and rho=0.5*

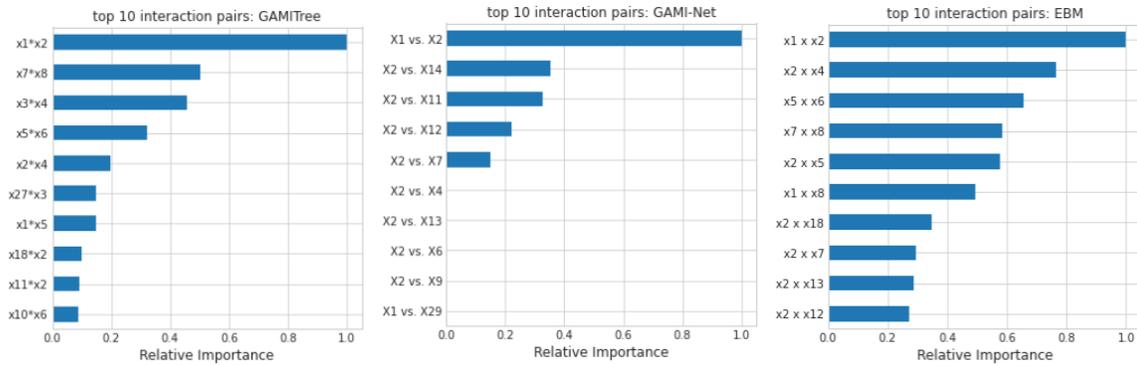

*Figure 26. Interaction importance for Model 3, 50K and rho=0.5*

In the binary case, the interaction patterns estimated by all algorithms are noisier and less accurate compared to continuous response case. This is due to the smaller signal-to-noise ratio for binary response. Figure 27 shows the interaction effects for Model 2, $\rho = 0.5$ and 50K data. The general patterns are more or less close to the truth in Figure 9, but much less accurate compared to continuous case. For example, for the interaction among $x_4$ and $x_5$, we didn't see the clear quadratic pattern for $x_5$ as in the continuous case. In addition, the scale (range of variation) is significantly smaller. However, with a larger sample size 500K, the model improves and the patterns become closer to the truth, as in Figure 28. In particular, we can see GAMI-Tree is closer to truth than EBM or GAMI-Net.

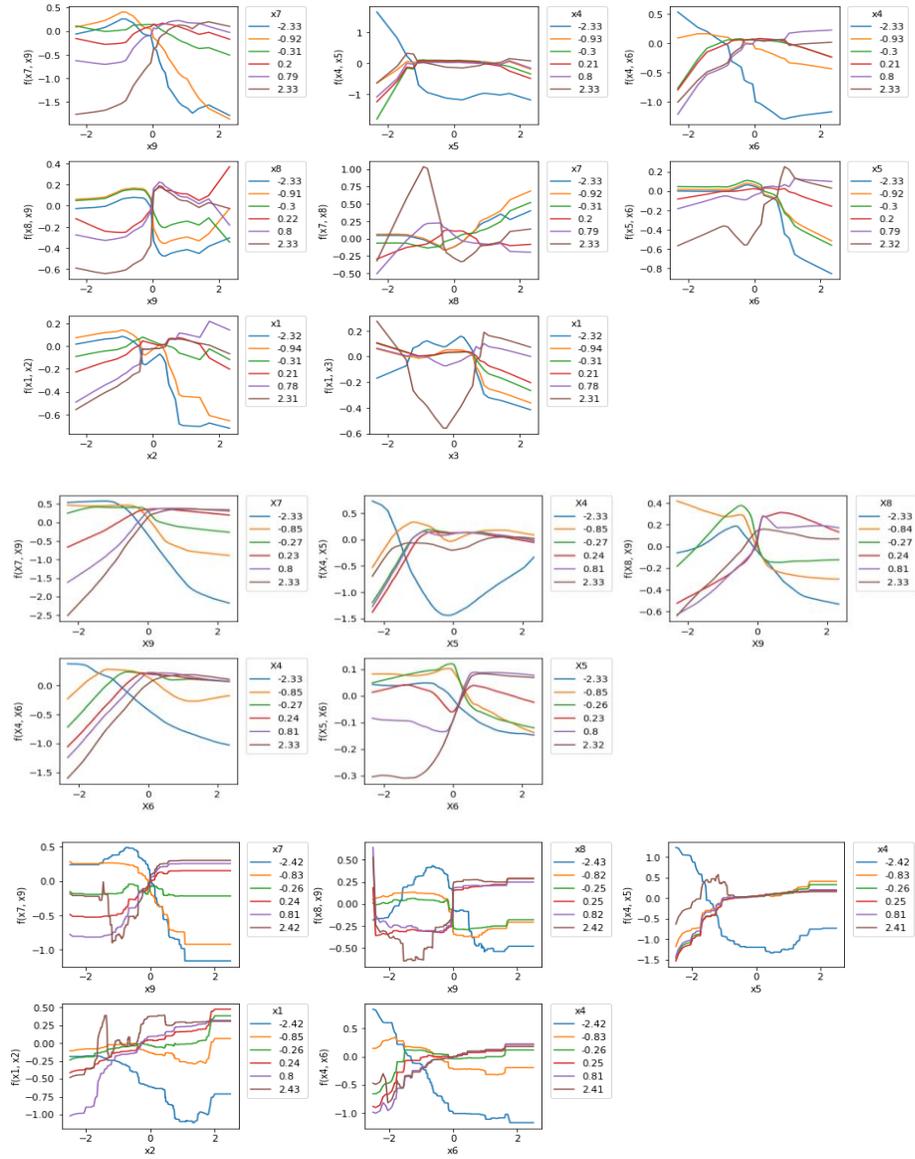

*Figure 27. Interaction effect plot for GAMI-Tree (top), GAMI-Net (middle), and EBM (bottom), for Model 2, 50K and rho=0.5*

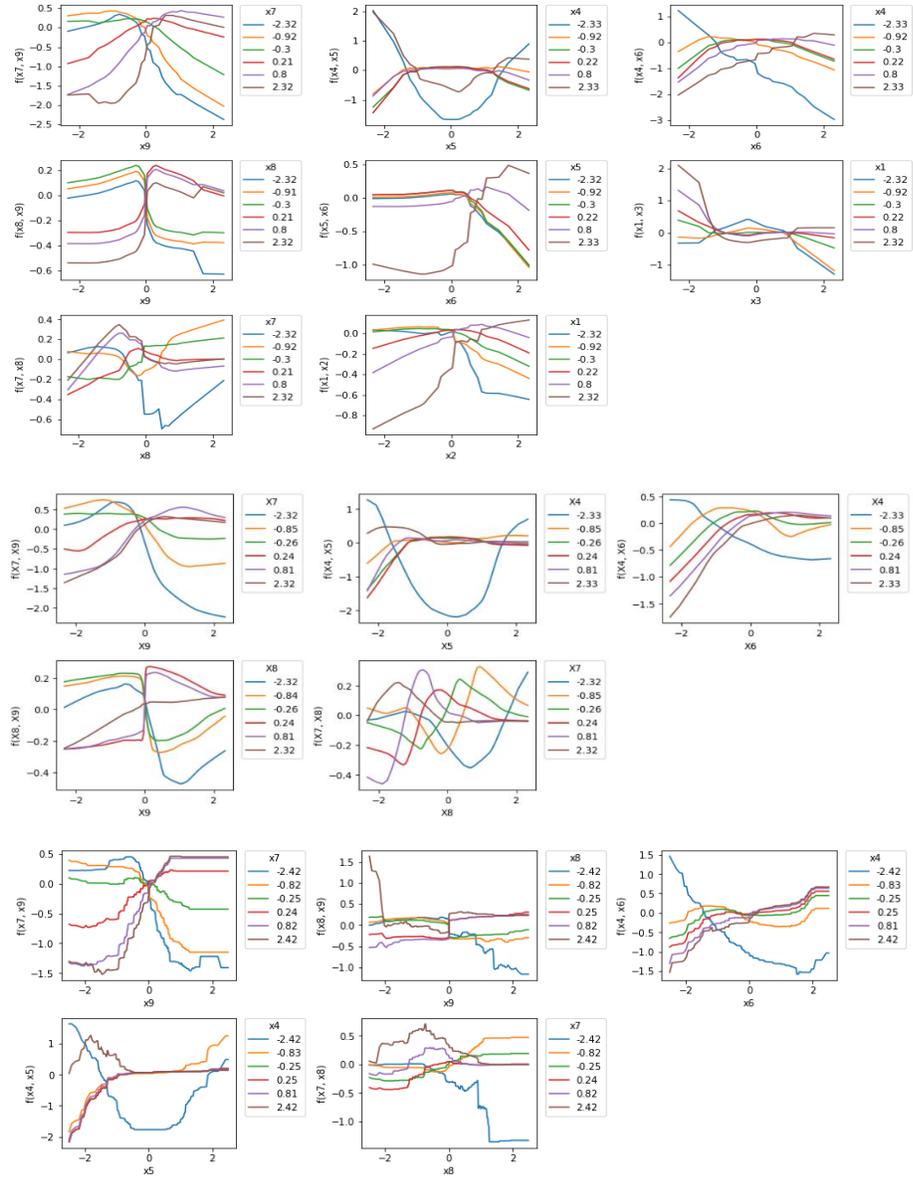

*Figure 28. Interaction effect plot for GAMI-Tree (top), GAMI-NET (middle), EBM (bottom), for Model 2, 500K and rho=0.5*